\pdfoutput=1
\documentclass{article}

\usepackage[margin=1in]{geometry}  
\usepackage{mathpazo}
\usepackage[backend=biber,style=alphabetic,natbib=true]{biblatex}
\usepackage[utf8]{inputenc} 
\usepackage[T1]{fontenc}    
\usepackage{hyperref}       
\usepackage{url}            
\usepackage{booktabs}       
\usepackage{amsfonts}       
\usepackage{nicefrac}       
\usepackage{microtype}      
\usepackage{longtable}  

\usepackage{amsmath}
\usepackage{comment}
\usepackage{array}
\usepackage{enumerate}
\usepackage{amsthm}
\usepackage[textsize=scriptsize]{todonotes}

\usepackage[utf8]{inputenc} 
\usepackage[T1]{fontenc}    
\usepackage{hyperref}       
\usepackage{url}            
\usepackage{booktabs}       
\usepackage{amsfonts}       
\usepackage{nicefrac}       
\usepackage{microtype}      
\usepackage{subcaption}
\usepackage{graphicx}
\usepackage{multirow}
\usepackage{dsfont}
\usepackage{algorithm}
\usepackage[algo2e]{algorithm2e} 
\usepackage{diagbox}
\usepackage{makecell}
\usepackage{longtable}

\usepackage{graphbox}

\newcommand{\eps}{\varepsilon}
\renewcommand{\epsilon}{\varepsilon}

\newcommand{\Breeds}{\textsc{Breeds}}
\newcommand{\allthree}{\textsc{Entity-13}}
\newcommand{\allfour}{\textsc{Entity-30}}
\newcommand{\living}{\textsc{Living-17}}
\newcommand{\nonliving}{\textsc{Non-living-26}}

\title{\Breeds: Benchmarks for Subpopulation Shift}

\author{
	Shibani Santurkar\footnotemark[1] \\
	MIT \\
	\texttt{shibani@mit.edu} \\
	\and
	Dimitris Tsipras\thanks{Equal contribution.} \\
	MIT \\
	\texttt{tsipras@mit.edu} 
	\and 
	Aleksander M\k{a}dry \\
	MIT \\
	\texttt{madry@mit.edu}  
}
\date{}

\begin{document}

\maketitle
\begin{abstract}
We develop a methodology for assessing the robustness of models  to 
subpopulation shift---specifically, their ability to generalize to novel 
data subpopulations that were not observed during training. 
Our approach leverages the class 
structure underlying existing datasets to control the data 
subpopulations that comprise the training and test distributions. This 
enables us to synthesize realistic distribution shifts whose 
sources can be precisely controlled and characterized, within existing
large-scale datasets. 
Applying this methodology to the ImageNet dataset, we create a suite of 
subpopulation shift benchmarks of varying granularity.
We then validate that the corresponding shifts are tractable by obtaining human
baselines for them.
Finally, we utilize these benchmarks to measure the sensitivity of 
standard model architectures as well as the effectiveness of 
off-the-shelf train-time robustness interventions. 
\footnote{Code and data available at 
\url{https://github.com/MadryLab/BREEDS-Benchmarks}.}

\end{abstract}

\section{Introduction}
\label{sec:intro}
Robustness to distribution shift has been the focus of a long line of
work in machine learning~\citep{schlimmer1986beyond,widmer1993effective,
kelly1999impact,shimodaira2000improving,sugiyama2007covariate,
quionero2009dataset,moreno2012unifying,sugiyama2012machine}.
At a high-level, the goal is to ensure that models perform well not only on 
unseen
samples from the datasets they are trained on, but also on the diverse set of
inputs they are likely to encounter in the real world.
However, building benchmarks for evaluating such robustness is 
challenging---it requires modeling realistic data variations in a  way that is 
well-defined, controllable, and easy to simulate.

Prior work in this context has focused on building benchmarks that capture 
distribution shifts caused by
natural or adversarial input
corruptions~\cite{szegedy2014intriguing,fawzi2015manitest,fawzi2016robustness,
    engstrom2019rotation,ford2019adversarial,hendrycks2019benchmarking,
    kang2019testing},
differences in data sources~\cite{saenko2010adapting,torralba2011unbiased,
    khosla2012undoing,tommasi2014testbed,recht2018imagenet},
and changes in the frequencies of data  
subpopulations~\cite{oren2019distributionally,sagawa2019distributionally}.
While each of these approaches captures a different source of real-world
distribution shift, we cannot expect any single benchmark to be 
comprehensive.
Thus, to obtain a holistic understanding of model robustness, we need 
to keep expanding our testbed to encompass more natural modes of variation.
In this work, we take another step in that direction by studying the following
question:

\begin{center}
	\emph{How well do models generalize to data subpopulations they have 
	not seen during training?}
\end{center}

\noindent
The notion of \emph{subpopulation shift} this question refers to is quite
pervasive.
After all, our training datasets will inevitably fail to perfectly 
capture the diversity of the real word.
Hence, during deployment, our models are bound to encounter unseen 
subpopulations---for instance, unexpected weather conditions in the 
self-driving car context or different diagnostic setups in medical applications.

\subsection*{Our contributions}
The goal of our work is to create large-scale subpopulation shift benchmarks
wherein the data subpopulations present during model training and evaluation
differ.
These benchmarks aim to assess how effectively models generalize beyond the
limited diversity of their training datasets---e.g., whether models can
recognize Dalmatians as ``dogs'' even when their training data for ``dogs''
 comprises only Poodles and Terriers.
We show how one can simulate such shifts, fairly naturally, \emph{within}
existing datasets, hence eliminating the need for (and the potential biases
introduced by) crafting synthetic transformations or collecting additional data.

\paragraph{\Breeds{} benchmarks.}
The crux of our approach is to leverage existing dataset labels and use them to
identify \emph{superclasses}---i.e., groups of semantically similar classes.
This allows us to construct classification tasks over such superclasses, and
repurpose the original dataset classes to be the subpopulations of interest.
This, in turn, enables us to induce a subpopulation shift by directly making
the subpopulations present in the training and test distributions 
disjoint.
By applying this methodology to the ImageNet 
dataset~\citep{deng2009imagenet}, we create a suite of subpopulation shift 
benchmarks of 
varying difficulty.
This involves modifying the existing ImageNet class
hierarchy---WordNet~\citep{miller1995wordnet}---to ensure that 
superclasses comprise visually coherent subpopulations.
We then conduct human studies to validate that the resulting \Breeds{}
benchmarks indeed capture meaningful subpopulation shifts.

\paragraph{Model robustness to subpopulation shift.} 
In order to demonstrate the utility of our benchmarks, we employ them 
to
evaluate the robustness of standard models to subpopulation
shift.
In general, we find that model performance drops significantly on the shifted
distribution---even when this shift does not significantly affect humans.
Still, models that are more accurate on the original distribution tend to also 
be more robust to these subpopulation shifts.
Moreover, adapting models to the shifted domain, by
retraining their last layer on data from this domain, only partially recovers the 
original 
model
performance.

\paragraph{Impact of robustness interventions.}
Finally, we examine whether various train-time interventions, designed to
decrease model sensitivity to synthetic data corruptions (e.g., $\ell_2$-bounded
perturbations) make models more robust to subpopulation shift.
We find that many of these methods offer small, yet non-trivial, 
improvements to
model robustness along this axis---at times, at the expense of performance on the
original distribution.
Often, these improvements become more pronounced after
retraining the last layer of the model on the shifted distribution.
In the context of adversarial training, our
findings are in line with recent work showing that the
resulting robust models 
often exhibit improved robustness to other data
corruptions~\citep{ford2019adversarial,kang2019testing,taori2020measuring}, and transfer
better to downstream
tasks~\citep{utrera2020adversarially,salman2020adversarially}.
Nonetheless, none of these interventions significantly alleviate
model sensitivity to subpopulation shift, indicating that the \Breeds{} 
benchmarks pose a challenge to current methods.

\section{Designing Benchmarks for Distribution Shift}
\label{sec:prior}
When constructing distribution shift benchmarks, the key design choice lies in 
specifying the \emph{target distribution} to be used during model evaluation.
This distribution is meant to be a realistic variation of the 
\emph{source distribution}, that was used for training.
Typically, studies focus on variations due to:
\begin{itemize}
    \item \emph{Data corruptions}: The target distribution is obtained by
        modifying inputs from the source distribution via a family of
        transformations that mimic real-world corruptions.
        Examples include natural or 
        adversarial forms of noise~\cite{fawzi2015manitest,fawzi2016robustness,
        engstrom2019rotation,hendrycks2019benchmarking,ford2019adversarial,kang2019testing,
        shankar2019image}.
    \item \emph{Differences in data sources}: Here, the target distribution is an
    independently collected dataset for the same 
    task~\cite{saenko2010adapting,torralba2011unbiased,tommasi2014testbed,
    beery2018recognition,recht2018imagenet}---for
    instance, using PASCAL VOC~\cite{everingham2010pascal} to evaluate
    ImageNet-trained classifiers~\cite{russakovsky2015imagenet}. The goal is to
    test whether models are overly reliant on the idiosyncrasies of the datasets
    they are trained
    on~\cite{ponce2006dataset,torralba2011unbiased}.
    \item \emph{Subpopulation shifts}: The source and target distributions 
    differ in terms of how well-represented each subpopulation is.
    Work in this area typically studies whether models perform 
    \emph{equally well} across
    all subpopulations from the perspective of
    reliability~\cite{meinshausen2015maximin, hu2016does,duchi2018learning,
    caldas2018leaf,oren2019distributionally,sagawa2019distributionally}
        or algorithmic 
    fairness~\citep{dwork2012fairness,kleinberg2017inherent,
    	jurgens2017incorporating, 
    	buolamwini2018gender,hashimoto2018fairness}.
\end{itemize} 

In general, a major challenge lies in ensuring that the distribution
shift between the source and target distributions (also referred to as 
\emph{domains}) is caused 
solely by the 
intended input variations.
External factors---which may arise when crafting synthetic
transformations or collecting new 
data---could skew the 
target distribution in different ways, making it hard to gauge model 
robustness to the exact distribution shift of interest.
For instance, recent work~\citep{engstrom2020identifying} demonstrates that
collecting a new dataset while aiming to match an existing one along a specific
metric (e.g., as in \citet{recht2018imagenet}) might result in a miscalibrated
dataset due to statistical bias.
In our study, we aim to limit such external influences by simulating
shifts within existing datasets, thus avoiding any input modifications.

\section{The \Breeds{} Methodology}
\label{sec:breeds}
\begin{figure}[!t]
	\centering
    \includegraphics[width=0.9\textwidth]{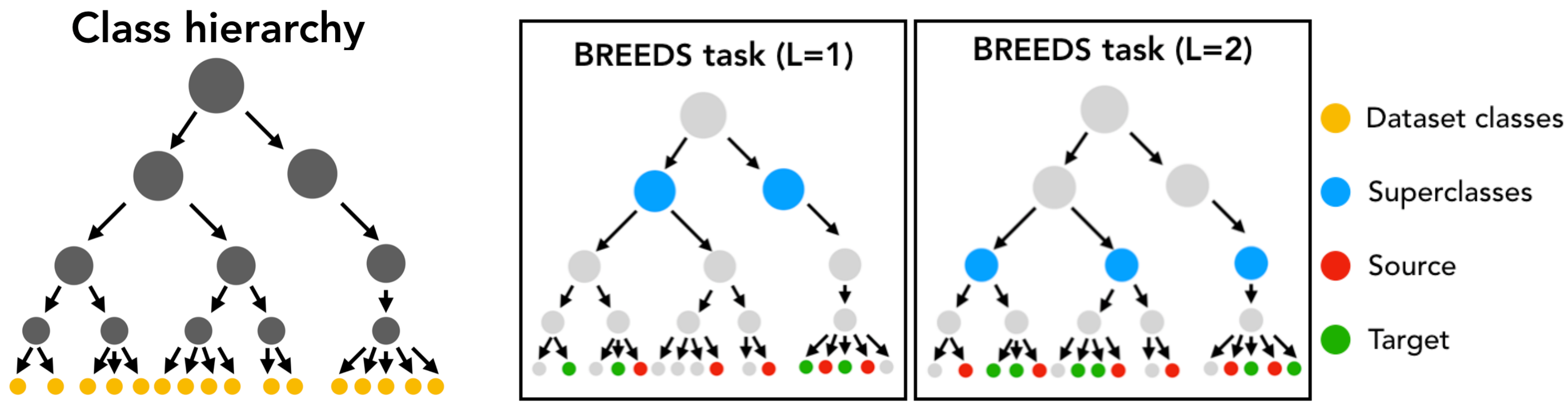}
    \caption{Illustration of our pipeline to create subpopulation shift
     benchmarks. Given a dataset, we first define superclasses by 
     grouping semantically similar classes together to form a hierarchy. This allows 
     us to treat the dataset labels as subpopulation annotations. Then, we 
     construct a \Breeds{} task of specified 
     granularity (i.e., depth in the hierarchy) by posing the 
     classification task 
     in terms of superclasses at that depth  and then partitioning their 
     respective 
     subpopulations into 
     the source and target domains.
 	}
    \label{fig:breeds}
\end{figure}

In this work, we focus on modeling a pertinent, yet relatively less studied,
form of subpopulation shift: one wherein the target distribution (used for 
testing) contains
subpopulations that are \emph{entirely} absent from the source distribution 
that the model was trained on.
To simulate such a shift, one needs to precisely control the data
subpopulations that comprise the source and target data distributions.
Our procedure for doing this comprises two stages that are outlined
below---see Figure~\ref{fig:breeds} for an illustration.

\paragraph{Devising subpopulation structure.}
Typical datasets do not contain 
annotations for individual subpopulations.
Since collecting such annotations would be challenging, we take an alternative
approach: we bootstrap the existing dataset labels to simulate 
subpopulations.
That is, we group semantically similar classes into broader
superclasses which, in turn, allows us to re-purpose existing class labels as 
the desired 
subpopulation annotations.
In fact, we can group classes in a hierarchical manner, obtaining superclasses
of different specificity.
As we will see in Section~\ref{sec:hierarchy}, large-scale benchmarks often
provide class hierarchies~\citep{everingham2010pascal,deng2009imagenet,
kuznetsova2018open} that aid such semantic grouping.

\paragraph{Simulating subpopulation shifts.}
Given a set of superclasses, we can define a classification task over them:
the inputs of each superclass correspond to pooling together the inputs
of its subclasses (i.e., the original dataset classes).
Within this setup, we can simulate subpopulation shift 
in a relatively straightforward manner.
Specifically, for each superclass, we split its subclasses into two
\emph{random} and \emph{disjoint} sets, and assign one of them to the source
and the other to the target domain.
Then, we can evaluate model robustness under subpopulation shift 
by simply training on the source domain and testing on the target domain.
Note that the classification task remains identical between
domains---both domains contain the same (super)classes but the subpopulations
that comprise each (super)class differ.
\footnote{Note that this methodology can be extended to simulate milder
subpopulation shifts where the source and target distributions overlap but the
relative subpopulation frequencies vary, similar to the setting of
\citet{oren2019distributionally}.}
Intuitively, this corresponds to using different dog breeds to represent the
class ``dog'' during training and testing---hence the name of our toolkit. 
\newline

\noindent
This methodology is quite general and can be applied to a variety of
setting to simulate realistic distribution shifts. 
Moreover, it has a number of additional benefits:
\begin{itemize}
    \item \textbf{Flexibility:} Different semantic groupings of a fixed set of
      classes lead to \Breeds{} tasks of varying granularity.
      For instance, by only grouping together classes that are quite similar one
      can reduce the severity of the subpopulation shift.
      Alternatively, one can consider broad superclasses, each having multiple
      subclasses, resulting in a more challenging benchmark.
    \item \textbf{Precise characterization:} The exact subpopulation shift
      between the source and target distribution is known.
      Since both domains are constructed from the same dataset, the impact
      of any external factors (e.g., differences in data collection pipelines) is
      minimized. (Note that such external factors can significantly impact the
      difficulty of the task~\cite{ponce2006dataset,
      torralba2011unbiased,engstrom2020identifying,tsipras2020from}.)
    \item \textbf{Symmetry:} Since subpopulations are split into the
      source and test domains randomly, we expect the resulting tasks to have
      comparable difficulty.
    \item \textbf{Reuse of existing datasets:} No additional data collection or
      annotation is required other than choosing the class grouping.
      This approach can thus be used to also re-purpose other existing
      large-scale datasets---even outside the image recognition context---with
      minimal effort and cost.
\end{itemize}

\section{Simulating Subpopulation Shifts Within ImageNet}
\label{sec:hierarchy}
We now describe in more detail how our methodology can be applied to
ImageNet~\citep{deng2009imagenet}---specifically, the ILSVRC2012
subset~\citep{russakovsky2015imagenet}---to create a suite of \Breeds{}
benchmarks.
ImageNet contains a large number of classes, making it particularly
well-suited for our purpose.

\subsection{Utilizing the ImageNet class hierarchy}
Recall that creating \Breeds{} tasks requires grouping together 
similar classes.
In the context of ImageNet, such a semantic grouping already 
exists---ImageNet classes
are a part of the WordNet hierarchy~\citep{miller1995wordnet}.
However, WordNet is not a hierarchy of objects but rather one of 
word
meanings.
Therefore, intermediate hierarchy nodes are not always well-suited
for object recognition due to:

\begin{itemize}
    \item \textbf{Abstract groupings:} WordNet nodes often correspond to
      abstract concepts, e.g., related to the functionality of an object.
      Children of such nodes might thus share little visual
      similarity---e.g., ``umbrella'' and ``roof'' are visually different,
      despite both being ``coverings''.
    \item \textbf{Non-uniform categorization:} The granularity of object
      categorization is vastly different across the WordNet hierarchy---e.g.,
      the subtree rooted at ``dog'' is 25-times larger than the one rooted at 
      ``cat''. 
      Hence, the depth of a node in this hierarchy does not always reflect
      the specificity of the corresponding object category.
    \item \textbf{Lack of tree structure:} Nodes in WordNet can have
      multiple parents\footnote{In programming languages, this is known as ``the
      diamond problem'' or ``the Deadly Diamond of
      Death''~\citep{martin1997java}.} and thus the resulting classification
      task would contain overlapping classes, making it inherently ambiguous.
\end{itemize}

\noindent
Due to these issues, we cannot directly use WordNet to identify  
superclasses that correspond to a well-calibrated classification task.
To illustrate this, we present some of the superclasses
constructed by applying clustering algorithms directly to the WordNet 
hierarchy~\cite{huh2016makes} in
Appendix Table~\ref{tab:problems}.
Even putting the issue of overlapping classes aside, a \Breeds{} task based on
these superclasses would induce a very skewed subpopulation shift across
classes---e.g., varying the types of ``bread'' is very different that
doing the same for different ``mammal'' species.

\paragraph{Calibrating WordNet for Visual Object Recognition.}
To better align the WordNet hierarchy with the task of object
recognition in general, and \Breeds{} benchmarks in particular, we manually 
modify it 
according to the following two principles.
First, nodes should be grouped together based on their visual characteristics,
rather than abstract relationships like functionality---e.g., we eliminate
nodes that do not convey visual information such as ``covering''.
Second, nodes of similar specificity should be at the same distance from the
root, irrespective of how detailed their categorization within WordNet is---for
instance, we placed ``dog'' at the same level as ``cat'' and ``flower'', even
though the ``dog'' sub-tree in WordNet is much larger.
Finally, we removed a number of ImageNet classes that did not naturally fit into
the hierarchy.
The resulting hierarchy, presented in Appendix~\ref{app:manual}, contains 
nodes of comparable granularity at the same level.
Moreover, as a result of this process, each node ends up having a single 
parent
and thus the resulting hierarchy is a tree.

\subsection{Creating \Breeds{} tasks}
\label{sec:tasks}
Once the modified version of the WordNet hierarchy is in place, \Breeds{} 
tasks can be
created in an automated manner.
Specifically, we first choose the desired granularity of the task by specifying 
the distance from the root (``entity'') and retrieving all superclasses at 
that distance in a top-down manner.
Each resulting superclass corresponds to a subtree of our hierarchy, with
ImageNet classes as its leaves.
Note that these superclasses are roughly of the same specificity, due to
our hierarchy restructuring process.
Then, we randomly sample a fixed number of subclasses for each 
superclass to produce a balanced dataset (omitting superclasses with an
insufficient number of subclasses).
Finally, as described in Section~\ref{sec:breeds}, we randomly split these
subclasses into the source and target domain.
\footnote{We also consider more benign or adversarial subpopulation
splits for these tasks in Appendix~\ref{app:goodbad}.}

\begin{figure}[!t]
	\centering
	\includegraphics[width=0.9\textwidth]{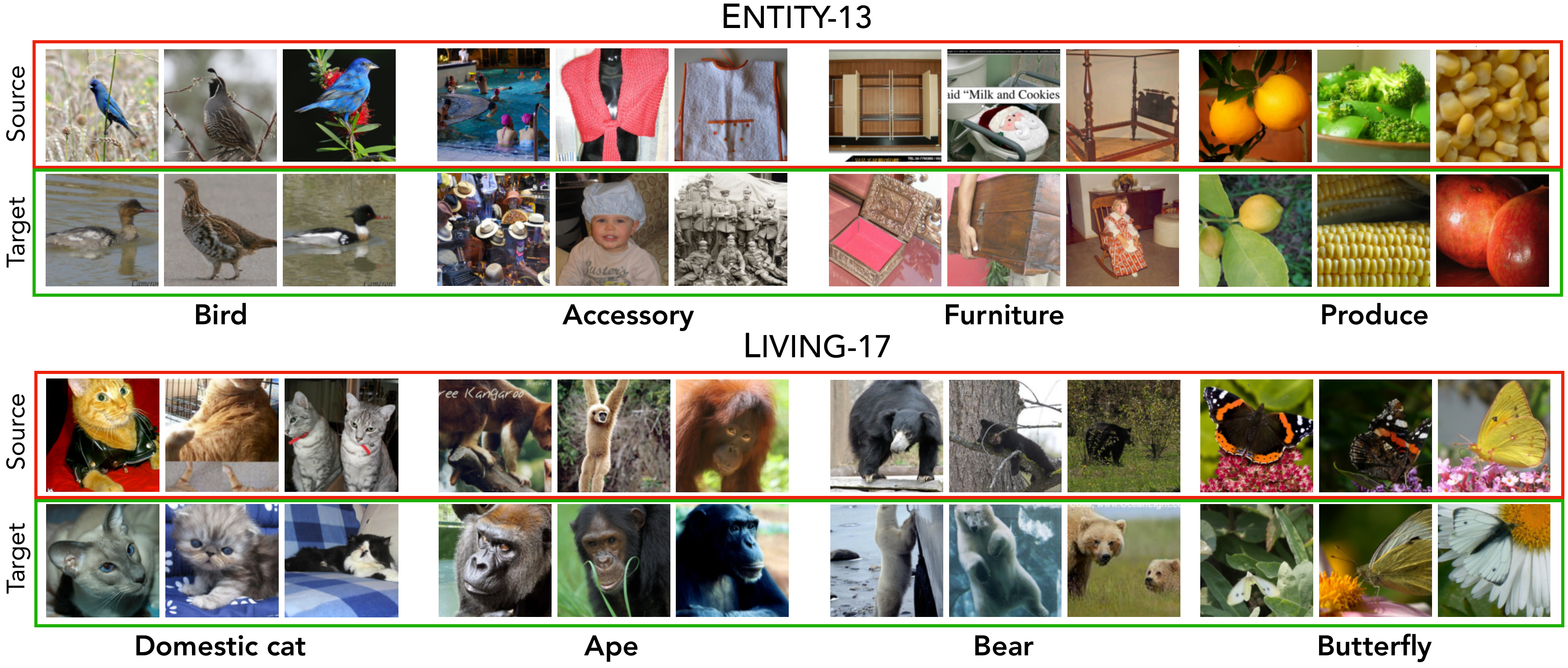}
	\caption{Sample images from random object categories for the \allthree{} 
        and \living{} tasks. For each task, the top and bottom row
        correspond to the source and target distributions respectively.}
	\label{fig:samples}
\end{figure}

For our analysis, we create four tasks, presented in 
Table~\ref{tab:benchmarks}, 
based on different levels/parts of the hierarchy.
To illustrate what the corresponding subpopulation shifts look like, we also 
present (random) image samples for a subset of 
the tasks in Figure~\ref{fig:samples}.
Note that while we focus on the tasks in Table~\ref{tab:benchmarks} in our 
study, our methodology readily enables us to create other 
variants of these tasks in an automated manner.

\begin{table}[!ht]
	\centering
	\begin{tabular}{lcccr}
		\toprule
		\textbf{Name} & \textbf{Subtree} & \textbf{Level} & 
		\textbf{Subpopulations} &
		\textbf{Examples} \\
		\midrule
		\allthree & ``entity'' (root) & 3 & 20 & ``mammal'', 
		``appliance'' \\
		\allfour & ``entity'' (root) & 4 & 8 & ``fruit'', ``carnivore''\\
		\living & ``living thing'' & 5 & 4 & ``ape'', ``bear'' \\
		\nonliving & ``non-living thing'' & 5 & 4 & ``fence'', ``ball''\\
		\bottomrule
	\end{tabular}
\vspace{1em}
	\caption{\Breeds{} benchmarks constructed using ImageNet.
	``Level'' indicates the depth of the 
  superclasses in the class hierarchy (task granularity). The number of
	``subpopulations'' (per superclass) is fixed across  
	superclasses to ensure a balanced 
	dataset.
	We can also construct specialized tasks, by focusing on subtrees in 
	the hierarchy, e.g., only living (\living{}) 
	or non-living (\nonliving{}) objects. 
	Datasets are named based on the root of the subtree and the resulting number
  of superclasses they end up containing.}
	\label{tab:benchmarks}
\end{table}

\paragraph{\Breeds{} benchmarks beyond ImageNet.} 
It is worth nothing that the methodology we described is not restricted to ImageNet and
can be readily applied to other datasets as well.
The only requirement is that we have access to a semantic grouping of the
dataset classes, which is the case for many popular 
vision datasets---e.g., 
CIFAR-100~\cite{krizhevsky2009learning}, 
Pascal-VOC~\cite{everingham2010pascal}, 
OpenImages~\cite{kuznetsova2018open}, 
COCO-Stuff~\cite{caesar2018cocostuff}.
Moreover, even when a class hierarchy is entirely
absent, the needed semantic class grouping can be manually
constructed with relatively little effort (proportional 
to the number of classes, not the number of datapoints).

\subsection{Calibrating \Breeds{} benchmarks via human studies}
\label{sec:humans}
For a distribution shift benchmark to be meaningful, it is essential that the source 
and target domains capture the same high-level task---otherwise generalizing
from one domain to the other would be impossible.
To ensure that this is the case for the \Breeds{} task, we assess how
significant the resulting distribution shifts are for human annotators
(crowd-sourced via MTurk).

\paragraph{Annotator task.} 
To obtain meaningful performance estimates, it is crucial that
annotators perform the task based {only} \emph{on the visual content of the
images}, without leveraging prior knowledge of the visual world.
To achieve this, we design the following annotation task.
First, annotators are shown images from the source domain, grouped by
superclass, without being aware of the superclass name (i.e., the object 
grouping it corresponds to).
Then, they are presented with images from the target domain and are asked to
assign each of them to one of the groups.
For simplicity, we only present two random superclasses at a time, effectively
simulating binary classification.
Annotator accuracy can be measured directly as the fraction of images that they
assign to the superclass to which these images belong.
We perform this experiment for each of the \Breeds{} tasks constructed in
Section~\ref{sec:tasks}.
As a point of comparison, we repeat this experiment without subpopulation 
shift
(test images are sampled from the source domain) and for the 
superclasses
constructed by~\citet{huh2016makes} using the WordNet hierarchy directly 
(cf.
Appendix~\ref{app:mturk}).

\begin{figure}[!ht]
	\centering
	\includegraphics[width=1\textwidth]{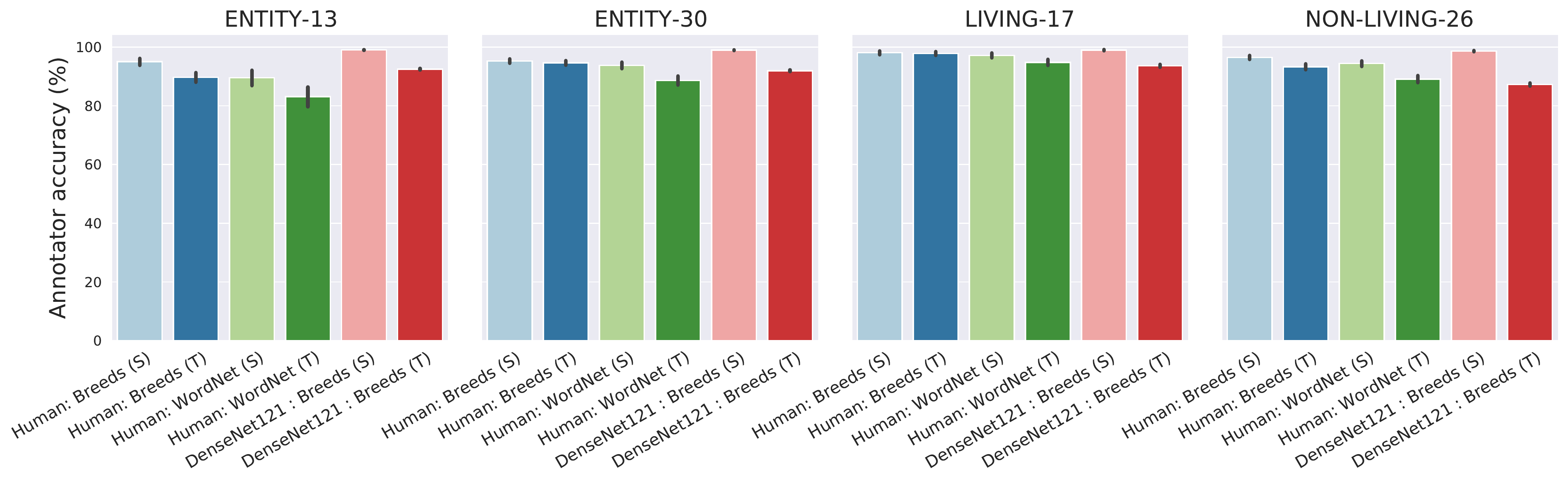}
    \caption{Human performance on (binary) \Breeds{} tasks.
		Annotators are provided with labeled 
        images from the source
		distribution for a \emph{pair of (undisclosed) superclasses}, and asked 
		to classify
	    samples from the target domain (`T') into one of the two groups. 
		As a baseline we also measure annotator performance without 
		subpopulation shift (i.e., on test images drawn from the source 
        domain, `S') and equivalent tasks created via the original WordNet
        hierarchy (cf. Appendix~\ref{app:mturk}).
		We can observe that across all tasks, annotators are fairly robust to 
		subpopulation shift. 
		Further, annotators consistently perform better on \Breeds{} task 
		compared to those based on WordNet directly---indicating that our 
		modified class hierarchy is indeed better calibrated for object 
		recognition.
		(We discuss model performance in Section~\ref{sec:eval}.)
	}
	\label{fig:acc_human}
\end{figure}

\paragraph{Human performance.} 
We find that, across all tasks, annotators perform well on unseen data from the
source domain, as expected.
More importantly, annotators also appear to be quite robust to subpopulation shift, 
experiencing only a small accuracy drop between the source and target 
domains (cf. Figure~\ref{fig:core}).
This is particularly prominent in the case of \allfour{} and \living{} where the
difference in source and target accuracy is within the confidence interval.
This indicates that the source and target domains are indeed perceptually 
similar for humans, making these benchmarks suitable for studying model 
robustness.
Finally, across all benchmarks, we observe that annotators perform better on 
\Breeds{} tasks, as compared to their WordNet equivalents---even on 
samples from the source domain.
This indicates that our modified ImageNet class hierarchy is indeed better 
aligned with the underlying visual object recognition task.

\section{Evaluating Model Performance under Subpopulation Shift}
\label{sec:eval}
We can now use our suite of \Breeds{} tasks as a testbed for assessing model
robustness to subpopulation shift as well as gauging the effectiveness of 
various
train-time robustness interventions. 
Specifics of the evaluation setup and additional experimental results 
are provided in Appendices~\ref{app:eval_setup}
and~\ref{app:res_eval}.

\subsection{Standard training}
\begin{figure}[!h]
	\centering
	\includegraphics[width=0.9\textwidth]{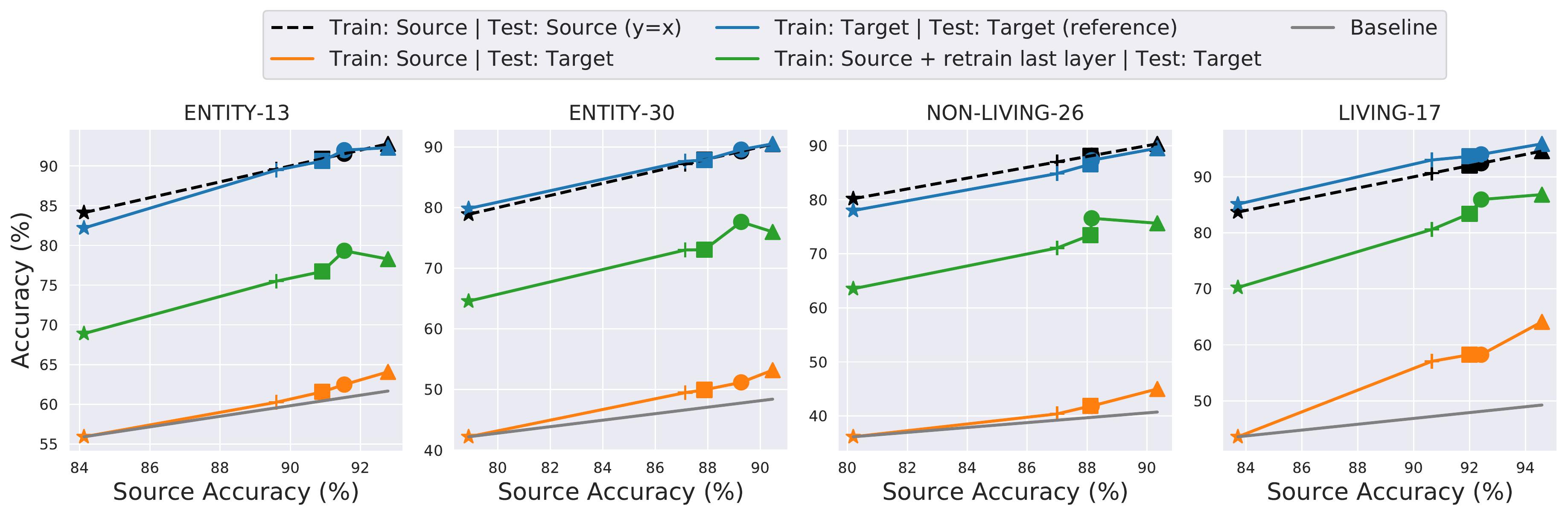}
	\caption{Robustness of standard models to \Breeds{} 
		subpopulation shifts. 
		For each of the four tasks, we plot the accuracy of different (source 
        domain-trained) model architectures (denoted by different symbols) on
        the target domain as a function of the source accuracy (which is
        typically high).
		We find that model accuracy drops significantly between domains
		(\emph{orange} vs.\ \emph{dashed} line). Still, models
		that are more accurate on the source domain seem to also be more 
		robust
		(the improvements exceed the baseline 
		(\emph{grey}) which would correspond to a constant accuracy drop
        across models, i.e., $\frac{source \ acc}{target \ acc}$ = 
		constant based on AlexNet).
		Moreover, the drop in model performance on the target domain can 
		be reduced by retraining
		the final model layer with data from that domain (\emph{green}). However, 
        a non-trivial drop persists compared to both the original source
        accuracy, and target accuracy of models trained directly (end-to-end)
        on the target domain (\emph{blue}).
	}
	\label{fig:core}
\end{figure}

We start by evaluating the performance of various model architectures trained
in the standard fashion: empirical risk minimization (ERM) on the source
distribution (cf. Appendix~\ref{app:models}).
While these models perform well on unseen inputs from the domain they are
trained on, i.e., they achieve high \emph{source accuracy}, they suffer a 
considerable
drop in accuracy under these subpopulation shifts---more than 30\% in most cases
(cf. Figure~\ref{fig:core}).
At the same time, models that are more \emph{accurate} on the 
source domain also appear to be more \emph{robust} to distribution shift.
Specifically, the fraction of source accuracy that is preserved in the target
domain is typically increasing with source accuracy. 
(If this were not the case, i.e., the accuracy of all models dropped by a 
constant fraction under distribution shift, the target accuracy would match 
the baseline in Figure~\ref{fig:core}.)
This indicates that, while models are quite brittle to subpopulation shift,
improvements in source accuracy \emph{do} correlate with models 
generalizing better to variations in testing conditions.
Note that model accuracies are not directly comparable across benchmarks, 
due to the presence of multiple conflating factors.
On one hand, more fine-grained tasks present a smaller subpopulation
shift (subclasses are semantically closer).
On the other hand, the number of classes and training inputs
per class changes significantly, making the task harder.

\paragraph{Models vs.\ Humans.} We compare 
the best performing model (DenseNet-121 in this case) to our previously 
obtained human baselines  in Figure~\ref{fig:acc_human}. To allow for a fair 
comparison, model accuracy is measured on pairwise superclass 
classification tasks (cf. Appendix~\ref{app:eval_setup}). We observe 
that models do exceedingly well on unseen samples from the source 
domain---significantly 
outperforming annotators under our task setup. At the same time, models 
also appear to be more brittle, performing worse than humans on the target
domain of these binary \Breeds{} tasks, despite their higher source accuracy.

\paragraph{Adapting models to the target domain.}
Finally, we focus on the intermediate data representations learned by these 
models, aiming to assess how suitable they are
for distinguishing classes in the target domain.
To assess this, we retrain the last (fully-connected) layer of models
trained on the source domain with data from the target domain. 
We find that the target accuracy of these models increases significantly after
retraining, indicating that the learned representations indeed generalize to
the target domain.
However, we cannot match the accuracy of models trained directly (end-to-end)
on the target domain---see Figure~\ref{fig:core}---demonstrating that there is
significant room for improvement.

\subsection{Robustness interventions}
\label{sec:intervene}
We now turn our attention to existing methods for decreasing 
model sensitivity to specific synthetic perturbations.
Our goal is to assess if these methods enhance model robustness to
subpopulation shift too.
Concretely, we consider the following families of 
interventions (cf. Appendix~\ref{app:robustness} for details):
\begin{itemize}
  \item \textbf{Adversarial training}: Enhances robustness to worst-case
  $\ell_p$-bounded perturbations (in our case $\ell_2$) by training models 
  against a projected gradient descent (PGD) adversary~\citep{madry2018towards}.
  \item \textbf{Stylized Training}: Encourages models to rely more on shape
    rather than texture by training them on a stylized version of 
    ImageNet~\cite{geirhos2018imagenettrained}.
  \item \textbf{Random noise}: Improves model robustness to data 
  corruptions by incorporating them as data augmentations during 
    training---we focus on Gaussian noise and Erase
    noise~\citep{zhong2020random}, i.e., randomly obfuscating a block of the
    image.
\end{itemize}

\noindent
Note that these methods can be viewed as ways of imposing a prior on the
features that the model relies on~\cite{heinze2017conditional,
    geirhos2018imagenettrained, engstrom2019learning}.
That is, by rendering certain features ineffective during training (e.g.,
texture) they incentivize the model to utilize alternative features
for its predictions (e.g., shape).
Since different families of features may correlate differently with class labels
in the target domain, the aforementioned interventions could significantly
impact model robustness to subpopulation shift.

\paragraph{Relative accuracy.}
To measure the impact of these interventions, we will focus on the
models' \emph{relative accuracy}---the ratio of target accuracy to source
accuracy.
This metric accounts for the fact that train-time interventions can impact model
accuracy on the source domain itself.
By measuring relative performance, we are able to compare different training
methods on an equal footing. 

\begin{figure}[!h]
	\centering
	\includegraphics[width=0.9\textwidth]{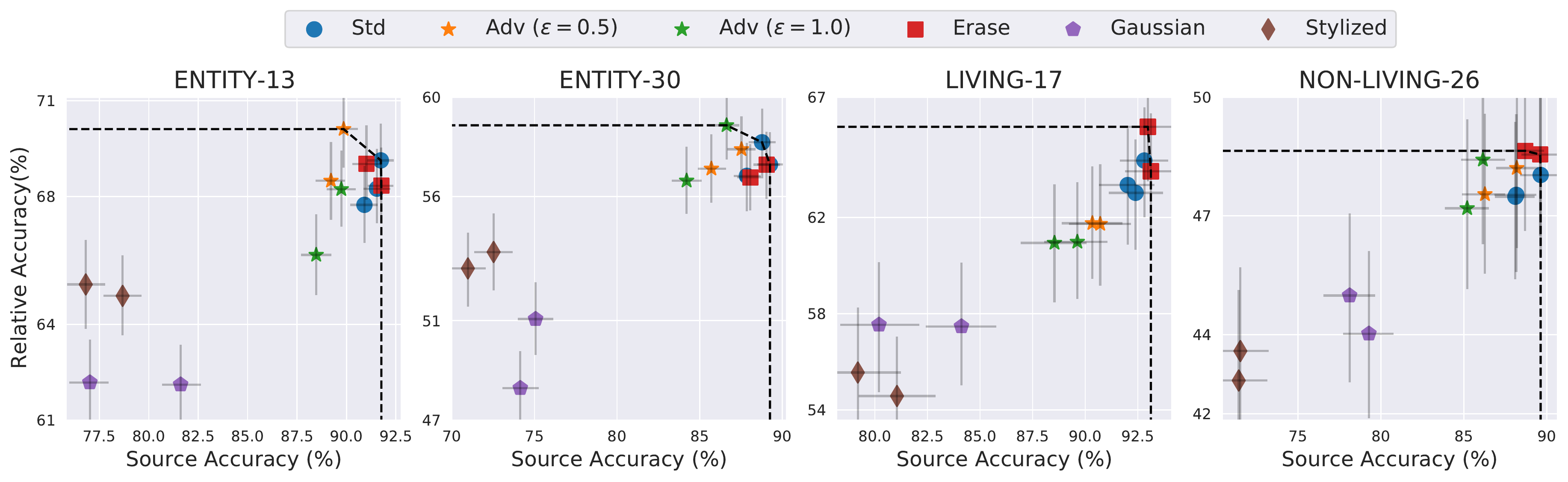}
	\caption{Effect of train-time interventions on model robustness to 
		subpopulation shift. We measure model performance in terms of 
		\emph{relative 
			accuracy}--i.e., the ratio between its target and source 
		accuracies. 
		This allows us to visualize the accuracy-robustness trade-off along with 
		the
		corresponding Pareto frontier (\emph{dashed}).
		(Also shown are 95\% confidence intervals computed via 
		bootstrapping.)
		We observe that some of these interventions do improve 
        model robustness to subpopulation shift by a small
        amount---specifically, erase noise and adversarial training---albeit
        sometimes at the cost of source accuracy.
	}
	\label{fig:intervene}
\end{figure}

We find that robustness interventions \emph{do} have a small,
yet non-trivial, impact on the robustness of a particular model architecture to
subpopulation shift---see Figure~\ref{fig:intervene}.
Specifically, for the case of adversarial training and erase noise, models often
retain a larger fraction of their accuracy to the target domain compared to
standard training, hence lying on the Pareto frontier of a robustness-accuracy
trade-off.
In fact, for some of the models trained with these interventions, the 
target accuracy is slightly higher than models obtained via standard training,
even without adjusting for their lower source accuracy (raw
accuracies for all methods are in Appendix~\ref{app:res_int}).
Nonetheless, it is important to note that none of these method offer
significant subpopulation robustness---relative accuracy is not
improved by more than a few percentage points.

\begin{figure}[!h]
	\centering
	\includegraphics[width=0.9\textwidth]{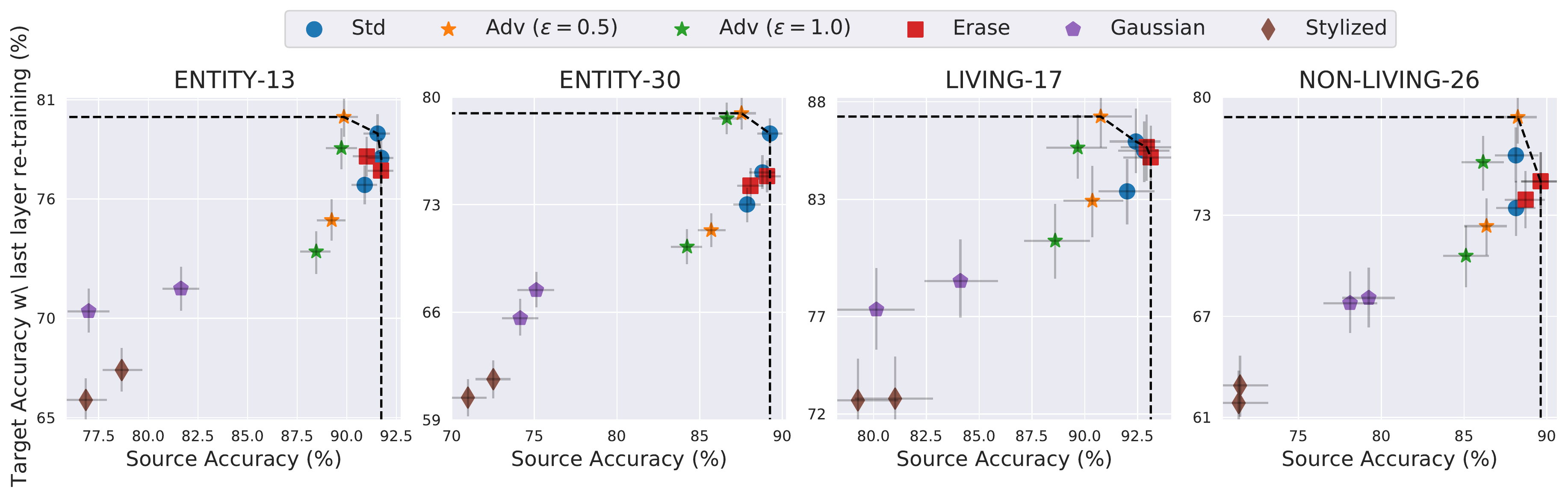}
	\caption{Target accuracy of models after
		they have been retrained (only the final linear layer) on data from the 
		target domain (with 95\% bootstrap confidence intervals).
		Models trained with robustness interventions often
	   have higher target accuracy than standard models post retraining.
   }
	\label{fig:intervene_ft}
\end{figure}

\paragraph{Adapting models to the target domain.}
The impact of these interventions is more pronounced if we consider  
the target accuracy of these models after their last layer has been retrained  on data from the target 
domain---see 
Figure~\ref{fig:intervene_ft}.
In particular, we observe that for adversarially robust models,  retraining
significantly boosts accuracy on the target domain---e.g., in the case of
\living{} it is almost comparable to the initial accuracy on the source domain.
This indicates that the feature priors imposed by these interventions
incentivize models to learn representations that generalize better to
similar domains---in line with recent results
of~\citet{utrera2020adversarially,salman2020adversarially}.
Moreover, we observe that models trained on the stylized version of these
datasets perform consistently worse, suggesting that texture might be an
important feature for these tasks, especially in the presence of subpopulation
shift.
Finally, note that we did not perform an exhaustive exploration of the
hyper-parameters used for these interventions (e.g., $\ell_2$-norm)---it is
possible that these results can be improved by additional tuning.
For instance, we would expect that we can tune the magnitude of the Gaussian
noise to achieve performance that is comparable to that of $\ell_2$-bounded
adversarial training~\citep{ford2019adversarial}.

\section{Additional Related Work}
In Section~\ref{sec:prior}, we discuss prior work that has directly focused on 
evaluating model robustness to distribution shift.  We now provide an 
overview of other related work and its connections to our methodology.

\paragraph{Distributional robustness.}
Distribution shifts that are small with respect to some $f$-divergence have been
studied in prior theoretical work~\citep{ben2013robust,duchi2016statistics,
esfahani2018data,namkoong2016stochastic}.
However, this notion of robustness is typically too pessimistic to capture
realistic data variations~\cite{hu2016does}.
Distributional robustness has also been connected to
causality~\citep{meinshausen2018causality}:
here, the typical approach is to inject spurious correlations into the dataset, 
and assess to what extent models rely on them for 
their predictions~\citep{heinze2017conditional,
arjovsky2019invariant,sagawa2019distributionally}.

\paragraph{Domain adaptation and transfer learning.} 
The goal here is to adapt models to the target domain with relatively few
samples from it~\citep{ben2007analysis, saenko2010adapting,ganin2014unsupervised,
courty2016optimal,gong2016domain,donahue2014decaf,razavian2014cnn}.
In domain adaptation, the task is the same in both domains,
while in transfer learning, the task itself could vary.
In a similar vein, the field of \emph{domain generalization} aims to generalize
to samples from a different domain (e.g., from ClipArt to photos) by training on
a number of explicitly annotated
domains~\citep{muandet2013domain,li2017deeper,peng2019moment}.

\paragraph{Zero-shot learning.}
Work in this domain focuses on learning to recognize previously unseen 
classes~\citep{lampert2009learning,xian2017zero}, typically described
via a semantic 
embedding~\citep{lampert2009learning,mikolov2013distributed,socher2013zero,frome2013devise,romera2015embarrassingly}.
This differs from our setup, where the focus is on  
generalization to unseen subpopulations for the \emph{same} set of classes.

\section{Conclusion}
In this work, we develop a methodology for constructing large-scale 
subpopulation shift benchmarks.
The motivation behind our \Breeds{} benchmarks is to
test if models can generalize beyond the limited diversity
of their training datasets---specifically, to novel data subpopulations.
A major advantage of our approach is its generality.
It can be applied to any dataset with a 
meaningful class structure---including tasks beyond classification 
(e.g., object detection) and domains other than computer vision 
(e.g., natural language processing).
Moreover, the subpopulation shifts are induced in a manner that is both
controlled and natural, without altering inputs synthetically or needing to 
collect new data.

We apply this approach to the ImageNet dataset to construct
benchmarks of varying difficulty.
We then demonstrate how these benchmarks can be 
used to assess model robustness and the efficacy of various train-time 
interventions.
Further, we obtain human baselines for these tasks to both put
model performance in context and validate that the corresponding 
subpopulation
shifts do not significantly affect humans.

Overall, our results indicate that existing models still have a long way to go
before they can fully tackle the BREEDS subpopulation shifts, even with 
robustness interventions.
We thus believe that our methodology provides a useful framework for studying
model robustness to distribution shift---an increasingly pertinent topic for
real-world deployments of machine learning models.

\section*{Acknowledgements}
We thank Andrew Ilyas and Sam Park for helpful discussions.

Work supported in part by the NSF grants CCF-1553428, CNS-1815221, the Google
PhD Fellowship, and the Microsoft Corporation. This material is based upon work supported by the Defense Advanced Research Projects Agency (DARPA) under Contract No. HR001120C0015.

Research was sponsored by the United States Air Force Research Laboratory and
was accomplished under Cooperative Agreement Number FA8750-19-2-1000. The views
and conclusions contained in this document are those of the authors and should
not be interpreted as representing the official policies, either expressed or
implied, of the United States Air Force or the U.S. Government. The U.S.
Government is authorized to reproduce and distribute reprints for Government
purposes notwithstanding any copyright notation herein.

\small 
\printbibliography
\clearpage
\normalsize
\appendix

\section{Experimental Setup}

\subsection{Dataset}
\label{app:datasets}
We perform our analysis on the ILSVRC2012
dataset~\citep{russakovsky2015imagenet}. This dataset contains a thousand
classes from the ImageNet dataset~\cite{deng2009imagenet} with an independently
collected validation set. 
The classes are part of the broader hierarchy, WordNet~\citep{miller1995wordnet}, 
through which
words are organized based on their semantic meaning.
We use this hierarchy as a starting point of our investigation but modify it
as described in Appendix~\ref{app:manual}.

For all the \Breeds{} superclass classification tasks, the train and validation
sets are obtained by aggregating the train and validation sets of the 
descendant ImageNet classes (i.e., subpopulations). 
Specifically, for a given subpopulation, the training and test splits from the 
original ImageNet dataset are used as is.

\subsection{WordNet issues}
\label{app:wordnet}

As discussed in Section~\ref{sec:hierarchy}, WordNet is a semantic rather than a
visual hierarchy. That is, object classes are arranged based on their meaning
rather than their visual appearance. Thus, using intermediate nodes for a
visual object recognition task is not straightforward.  To illustrate this, we
examine a sample superclass grouping created by~\citet{huh2016makes} via
automated bottom-up clustering in Table~\ref{tab:problems}.

\begin{table}[htp]
    
\makebox[\textwidth]{
\begin{tabular}{lp{.7\textwidth}}
    \toprule
        \textbf{Superclass} & \textbf{Random ImageNet classes} \\

\midrule \textbf{instrumentality} & fire engine, basketball, electric fan, wok, thresher, horse cart, harvester, balloon, racket, can opener, carton, gong, unicycle, toilet seat, carousel, hard disc, cello, mousetrap, neck brace, barrel \\ 
\midrule \textbf{man-made structure} & beacon, yurt, picket fence, barbershop, fountain, steel arch bridge, library, cinema, stone wall, worm fence, palace, suspension bridge, planetarium, monastery, mountain tent, sliding door, dam, bakery, megalith, pedestal \\ 
\midrule \textbf{covering} & window shade, vestment, running shoe, diaper, sweatshirt, breastplate, shower curtain, shoji, miniskirt, knee pad, apron, pajama, military uniform, theater curtain, jersey, football helmet, book jacket, bow tie, suit, cloak \\ 
\midrule \textbf{commodity} & espresso maker, maillot, iron, bath towel, lab coat, bow tie, washer, jersey, mask, waffle iron, mortarboard, diaper, bolo tie, seat belt, cowboy hat, wig, knee pad, vacuum, microwave, abaya \\ 
\midrule \textbf{organism} & thunder snake, stingray, grasshopper, barracouta, Newfoundland, Mexican hairless, Welsh springer spaniel, bluetick, golden retriever, keeshond, African chameleon, jacamar, water snake, Staffordshire bullterrier, Old English sheepdog, pelican, sea lion, wire-haired fox terrier, flamingo, green mamba \\ 
\midrule \textbf{produce} & spaghetti squash, fig, cardoon, mashed potato, pineapple, zucchini, broccoli, cauliflower, butternut squash, custard apple, pomegranate, strawberry, Granny Smith, lemon, head cabbage, artichoke, cucumber, banana, bell pepper, acorn squash \\

    \bottomrule
\end{tabular}
}

    \vspace{1em}
    \caption{Superclasses constructed by~\citet{huh2016makes} via
    bottom-up clustering of WordNet to obtain 36 superclasses---for
    brevity, we only show superclasses with at least 20 ImageNet classes each.}
    \label{tab:problems}
\end{table}

First, we can notice that these superclasses have vastly different
granularities.
For instance, ``organism'' contains the entire animal kingdom, hence being much
broader than ``produce''.
Moreover, ``covering'' is rather abstract class, and hence its subclasses often
share little visual similarity (e.g., ``window shade'', ``pajama'').
Finally, due to the abstract nature of these superclasses, a large number of
subclasses overlap---``covering'' and ``commodity'' share 49 ImageNet 
descendants.

\clearpage
\subsection{Manual calibration}
\label{app:method}
In order to allow for efficient and automated creation of superclasses that are
suitable for visual recognition, we modified the WordNet hierarchy by applying
the following operations:
\begin{itemize}
    \item \emph{Collapse node}: Delete a node from the hierarchy and add edges
        from each parent to each child. Allows us to remove redundant or overly
        specific categorization while preserving the overall structure.
    \item \emph{Insert node above}: Add a dummy parent to push a node further
        down the hierarchy. Allows us to ensure that nodes of similar granularity are at 
        the same level.
    \item \emph{Delete node}: Remove a node and all of its edges. Used to
        remove abstract nodes that do not reveal visual characteristics.
    \item \emph{Add edge}: Connect a node to a parent. Used to reassign the
        children of nodes deleted by the operation above.
\end{itemize}
We manually examined the hierarchy and implemented these actions in order to
produce superclasses that are calibrated for classification.
The principles we followed are outlined in Section~\ref{sec:hierarchy} while
the full hierarchy can be explored using the notebooks provided with the 
hierarchy.\footnote{\url{https://github.com/MadryLab/BREEDS-Benchmarks}}

\subsection{Resulting hierarchy}
\label{app:manual}

The parameters for constructing the \Breeds{} benchmarks (hierarchy level,
number of subclasses, and tree root) are given in Table~\ref{tab:benchmarks}.
The resulting tasks---obtained by sampling disjoint ImageNet classes (i.e., 
subpopulations) for the
source and target domain---are shown in
Tables~\ref{tab:three},~\ref{tab:four},~\ref{tab:living},
and~\ref{tab:nonliving}.
Recall that for each superclass we randomly sample a fixed number of 
subclasses per superclass to ensure that the dataset is approximately 
balanced.
\clearpage
\begin{longtable}{lp{.38\textwidth}p{.38\textwidth}}
	\toprule
	\textbf{Superclass} & \textbf{Source} & \textbf{Target} \\
	
	\midrule \textbf{garment} & trench coat, abaya, gown, poncho, military 
	uniform, jersey, cloak, bikini, miniskirt, swimming trunks & lab coat, 
	brassiere, hoopskirt, cardigan, pajama, academic gown, apron, diaper, 
	sweatshirt, sarong \\ 
	\midrule \textbf{bird} & African grey, bee eater, coucal, American coot, 
	indigo bunting, king penguin, spoonbill, limpkin, quail, kite & prairie 
	chicken, red-breasted merganser, albatross, water ouzel, goose, 
	oystercatcher, American egret, hen, lorikeet, ruffed grouse \\ 
	\midrule \textbf{reptile} & Gila monster, agama, triceratops, African 
	chameleon, thunder snake, Indian cobra, green snake, mud turtle, water 
	snake, loggerhead & sidewinder, leatherback turtle, boa constrictor, garter 
	snake, terrapin, box turtle, ringneck snake, rock python, American 
	chameleon, green lizard \\ 
	\midrule \textbf{arthropod} & rock crab, black and gold garden spider, 
	tiger beetle, black widow, barn spider, leafhopper, ground beetle, fiddler 
	crab, bee, walking stick & cabbage butterfly, admiral, lacewing, trilobite, 
	sulphur butterfly, cicada, garden spider, leaf beetle, long-horned beetle, fly 
	\\ 
	\midrule \textbf{mammal} & Siamese cat, ibex, tiger, hippopotamus, 
	Norwegian elkhound, dugong, colobus, Samoyed, Persian cat, Irish 
	wolfhound & English setter, llama, lesser panda, armadillo, indri, giant 
	schnauzer, pug, Doberman, American Staffordshire terrier, beagle \\ 
	\midrule \textbf{accessory} & bib, feather boa, stole, plastic bag, bathing 
	cap, cowboy boot, necklace, crash helmet, gasmask, maillot & hair slide, 
	umbrella, pickelhaube, mitten, sombrero, shower cap, sock, running shoe, 
	mortarboard, handkerchief \\ 
	\midrule \textbf{craft} & catamaran, speedboat, fireboat, yawl, airliner, 
	container ship, liner, trimaran, space shuttle, aircraft carrier & schooner, 
	gondola, canoe, wreck, warplane, balloon, submarine, pirate, lifeboat, 
	airship \\ 
	\midrule \textbf{equipment} & volleyball, notebook, basketball, hand-held 
	computer, tripod, projector, barbell, monitor, croquet ball, balance beam & 
	cassette player, snorkel, horizontal bar, soccer ball, racket, baseball, 
	joystick, microphone, tape player, reflex camera \\ 
	\midrule \textbf{furniture} & wardrobe, toilet seat, file, mosquito net, 
	four-poster, bassinet, chiffonier, folding chair, fire screen, shoji & studio 
	couch, throne, crib, rocking chair, dining table, park bench, chest, window 
	screen, medicine chest, barber chair \\ 
	\midrule \textbf{instrument} & upright, padlock, lighter, steel drum, parking 
	meter, cleaver, syringe, abacus, scale, corkscrew & maraca, saltshaker, 
	magnetic compass, accordion, digital clock, screw, can opener, odometer, 
	organ, screwdriver \\ 
	\midrule \textbf{man-made structure} & castle, bell cote, fountain, 
	planetarium, traffic light, breakwater, cliff dwelling, monastery, prison, 
	water tower & suspension bridge, worm fence, turnstile, tile roof, beacon, 
	street sign, maze, chainlink fence, bakery, drilling platform \\ 
	\midrule \textbf{wheeled vehicle} & snowplow, trailer truck, racer, shopping 
	cart, unicycle, motor scooter, passenger car, minibus, jeep, recreational 
	vehicle & jinrikisha, golfcart, tow truck, ambulance, bullet train, fire engine, 
	horse cart, streetcar, tank, Model T \\ 
	\midrule \textbf{produce} & broccoli, corn, orange, cucumber, spaghetti 
	squash, butternut squash, acorn squash, cauliflower, bell pepper, fig & 
	pomegranate, mushroom, strawberry, lemon, head cabbage, Granny Smith, 
	hip, ear, banana, artichoke \\

	\bottomrule
 \caption{Superclasses used for the \allthree{} task, along with the 
	corresponding 
	    subpopulations that comprise the source and target domains.}
	    \label{tab:three}
\end{longtable}

\clearpage
\begin{longtable}{lp{.38\textwidth}p{.38\textwidth}}
    \toprule
        \textbf{Superclass} & \textbf{Source} & \textbf{Target} \\

\midrule \textbf{serpentes} & green mamba, king snake, garter snake, thunder snake & boa constrictor, green snake, ringneck snake, rock python \\ 
\midrule \textbf{passerine} & goldfinch, brambling, water ouzel, chickadee & magpie, house finch, indigo bunting, bulbul \\ 
\midrule \textbf{saurian} & alligator lizard, Gila monster, American chameleon, green lizard & Komodo dragon, African chameleon, agama, banded gecko \\ 
\midrule \textbf{arachnid} & harvestman, barn spider, scorpion, black widow & wolf spider, black and gold garden spider, tick, tarantula \\ 
\midrule \textbf{aquatic bird} & albatross, red-backed sandpiper, crane, white stork & goose, dowitcher, limpkin, drake \\ 
\midrule \textbf{crustacean} & crayfish, spiny lobster, hermit crab, Dungeness crab & king crab, rock crab, American lobster, fiddler crab \\ 
\midrule \textbf{carnivore} & Italian greyhound, black-footed ferret, Bedlington terrier, basenji & flat-coated retriever, otterhound, Shih-Tzu, Boston bull \\ 
\midrule \textbf{insect} & lacewing, fly, grasshopper, sulphur butterfly & long-horned beetle, leafhopper, dung beetle, admiral \\ 
\midrule \textbf{ungulate} & llama, gazelle, zebra, ox & hog, hippopotamus, hartebeest, warthog \\ 
\midrule \textbf{primate} & baboon, howler monkey, Madagascar cat, chimpanzee & siamang, indri, capuchin, patas \\ 
\midrule \textbf{bony fish} & coho, tench, lionfish, rock beauty & sturgeon, puffer, eel, gar \\ 
\midrule \textbf{barrier} & breakwater, picket fence, turnstile, bannister & chainlink fence, stone wall, dam, worm fence \\ 
\midrule \textbf{building} & bookshop, castle, mosque, butcher shop & grocery store, toyshop, palace, beacon \\ 
\midrule \textbf{electronic equipment} & printer, pay-phone, microphone, computer keyboard & modem, cassette player, monitor, dial telephone \\ 
\midrule \textbf{footwear} & clog, Loafer, maillot, running shoe & sandal, knee pad, cowboy boot, Christmas stocking \\ 
\midrule \textbf{garment} & academic gown, apron, miniskirt, fur coat & jean, vestment, sarong, swimming trunks \\ 
\midrule \textbf{headdress} & pickelhaube, hair slide, shower cap, bonnet & bathing cap, cowboy hat, bearskin, crash helmet \\ 
\midrule \textbf{home appliance} & washer, microwave, Crock Pot, vacuum & toaster, espresso maker, space heater, dishwasher \\ 
\midrule \textbf{kitchen utensil} & measuring cup, cleaver, coffeepot, spatula & frying pan, cocktail shaker, tray, caldron \\ 
\midrule \textbf{measuring instrument} & digital watch, analog clock, parking meter, magnetic compass & barometer, wall clock, hourglass, digital clock \\ 
\midrule \textbf{motor vehicle} & limousine, school bus, moped, convertible & trailer truck, beach wagon, police van, garbage truck \\ 
\midrule \textbf{musical instrument} & French horn, maraca, grand piano, upright & acoustic guitar, organ, electric guitar, violin \\ 
\midrule \textbf{neckwear} & feather boa, neck brace, bib, Windsor tie & necklace, stole, bow tie, bolo tie \\ 
\midrule \textbf{sports equipment} & ski, dumbbell, croquet ball, racket & rugby ball, balance beam, horizontal bar, tennis ball \\ 
\midrule \textbf{tableware} & mixing bowl, water jug, beer glass, water bottle & goblet, wine bottle, coffee mug, plate \\ 
\midrule \textbf{tool} & quill, combination lock, padlock, screw & fountain pen, screwdriver, shovel, torch \\ 
\midrule \textbf{vessel} & container ship, lifeboat, aircraft carrier, trimaran & liner, wreck, catamaran, yawl \\ 
\midrule \textbf{dish} & potpie, mashed potato, pizza, cheeseburger & burrito, hot pot, meat loaf, hotdog \\ 
\midrule \textbf{vegetable} & zucchini, cucumber, butternut squash, artichoke & cauliflower, spaghetti squash, acorn squash, cardoon \\ 
\midrule \textbf{fruit} & strawberry, pineapple, jackfruit, Granny Smith & buckeye, corn, ear, acorn \\

    \bottomrule
    \caption{Superclasses used for the \allfour{} task, along with the 
    corresponding 
	subpopulations that comprise the source and target domains.}
\label{tab:four}
\end{longtable}

\clearpage
\begin{longtable}{lp{.38\textwidth}p{.38\textwidth}}
    \toprule
        \textbf{Superclass} & \textbf{Source} & \textbf{Target} \\

\midrule \textbf{salamander} & eft, axolotl & common newt, spotted salamander \\ 
\midrule \textbf{turtle} & box turtle, leatherback turtle & loggerhead, mud turtle \\ 
\midrule \textbf{lizard} & whiptail, alligator lizard & African chameleon, banded gecko \\ 
\midrule \textbf{snake} & night snake, garter snake & sea snake, boa constrictor \\ 
\midrule \textbf{spider} & tarantula, black and gold garden spider & garden spider, wolf spider \\ 
\midrule \textbf{grouse} & ptarmigan, prairie chicken & ruffed grouse, black grouse \\ 
\midrule \textbf{parrot} & macaw, lorikeet & African grey, sulphur-crested cockatoo \\ 
\midrule \textbf{crab} & Dungeness crab, fiddler crab & rock crab, king crab \\ 
\midrule \textbf{dog} & bloodhound, Pekinese & Great Pyrenees, papillon \\ 
\midrule \textbf{wolf} & coyote, red wolf & white wolf, timber wolf \\ 
\midrule \textbf{fox} & grey fox, Arctic fox & red fox, kit fox \\ 
\midrule \textbf{domestic cat} & tiger cat, Egyptian cat & Persian cat, Siamese cat \\ 
\midrule \textbf{bear} & sloth bear, American black bear & ice bear, brown bear \\ 
\midrule \textbf{beetle} & dung beetle, rhinoceros beetle & ground beetle, long-horned beetle \\ 
\midrule \textbf{butterfly} & sulphur butterfly, admiral & cabbage butterfly, ringlet \\ 
\midrule \textbf{ape} & gibbon, orangutan & gorilla, chimpanzee \\ 
\midrule \textbf{monkey} & marmoset, titi & spider monkey, howler monkey \\

    \bottomrule
    \caption{Superclasses used for the \living{} task, along with the 
    corresponding 
	subpopulations that comprise the source and target domains.}
\label{tab:living}
\end{longtable}

\clearpage
\begin{longtable}{lp{.38\textwidth}p{.38\textwidth}}
    \toprule
        \textbf{Superclass} & \textbf{Source} & \textbf{Target} \\

\midrule \textbf{bag} & plastic bag, purse & mailbag, backpack \\ 
\midrule \textbf{ball} & volleyball, punching bag & ping-pong ball, soccer ball \\ 
\midrule \textbf{boat} & gondola, trimaran & catamaran, canoe \\ 
\midrule \textbf{body armor} & bulletproof vest, breastplate & chain mail, cuirass \\ 
\midrule \textbf{bottle} & pop bottle, beer bottle & wine bottle, water bottle \\ 
\midrule \textbf{bus} & trolleybus, minibus & school bus, recreational vehicle \\ 
\midrule \textbf{car} & racer, Model T & police van, ambulance \\ 
\midrule \textbf{chair} & folding chair, throne & rocking chair, barber chair \\ 
\midrule \textbf{coat} & lab coat, fur coat & kimono, vestment \\ 
\midrule \textbf{digital computer} & laptop, desktop computer & notebook, hand-held computer \\ 
\midrule \textbf{dwelling} & palace, monastery & mobile home, yurt \\ 
\midrule \textbf{fence} & worm fence, chainlink fence & stone wall, picket fence \\ 
\midrule \textbf{hat} & bearskin, bonnet & sombrero, cowboy hat \\ 
\midrule \textbf{keyboard instrument} & grand piano, organ & upright, accordion \\ 
\midrule \textbf{mercantile establishment} & butcher shop, barbershop & shoe shop, grocery store \\ 
\midrule \textbf{outbuilding} & greenhouse, apiary & barn, boathouse \\ 
\midrule \textbf{percussion instrument} & steel drum, marimba & drum, gong \\ 
\midrule \textbf{pot} & teapot, Dutch oven & coffeepot, caldron \\ 
\midrule \textbf{roof} & dome, vault & thatch, tile roof \\ 
\midrule \textbf{ship} & schooner, pirate & aircraft carrier, liner \\ 
\midrule \textbf{skirt} & hoopskirt, miniskirt & overskirt, sarong \\ 
\midrule \textbf{stringed instrument} & electric guitar, banjo & violin, acoustic guitar \\ 
\midrule \textbf{timepiece} & digital watch, stopwatch & parking meter, digital clock \\ 
\midrule \textbf{truck} & fire engine, pickup & tractor, forklift \\ 
\midrule \textbf{wind instrument} & oboe, sax & flute, bassoon \\ 
\midrule \textbf{squash} & spaghetti squash, acorn squash & zucchini, butternut squash \\

    \bottomrule
    \caption{Superclasses used for the \nonliving{} task, along with the 
corresponding 
    subpopulations that comprise the source and target domains.}
    \label{tab:nonliving}
\end{longtable}

\clearpage
\subsection{Annotator task}
\label{app:mturk}
\begin{figure}[htp]
    \centering
    \includegraphics[width=0.78\textwidth]{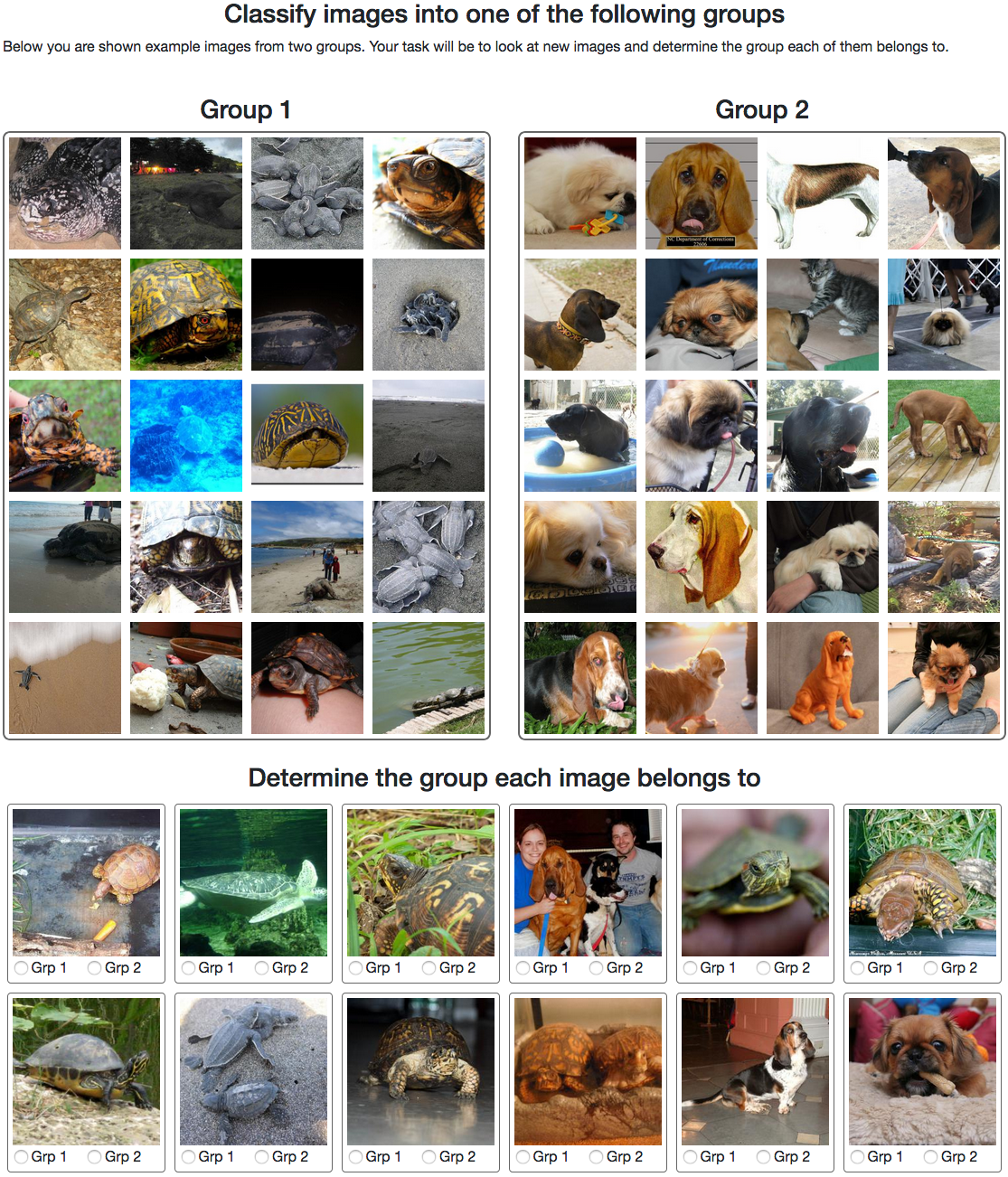}
    \caption{Sample MTurk annotation task to obtain human baselines for \Breeds{} 
    benchmarks.}
    \label{fig:screenshot}
\end{figure}

As described in Section~\ref{sec:humans}, the goal of our human studies is to
understand whether humans can classify images into superclasses even without
knowing the semantic grouping.
Thus, the task involved showing annotators two groups of images, each sampled
from the source domain of a random superclass.
Then, annotators were shown a new set of images from the target domain (or the
source domain in the case of control) and were asked to assign each of them into
one of the two groups. A screenshot of an (random) instance of our annotator
task is shown in Figure~\ref{fig:screenshot}.

Each task contained 20 images from the source domain of each superclass and 12
images for annotators to classify (the images where rescaled and center-cropped
to size $224\times 224$ to match the input size use for model predictions).
The two superclasses were randomly permuted at load time.
To ensure good concentration of our accuracy estimates, for every superclass, 
we 
performed binary classification tasks w.r.t. 3 other (randomly chosen) superclasses.
Further, we used 3 annotators per task. and annotators were compensated \$0.15 
per task.

\paragraph{Comparing with the original hierarchy.} In order to compare our
superclasses with those obtained by \citet{huh2016makes} via WordNet
clustering,\footnote{\url{https://github.com/minyoungg/wmigftl/tree/master/label_sets/hierarchy}}
we need to define a correspondence between them.
To do so, for each of our tasks, we selected the clustering (either top-down or
bottom-up) that had the closest number of superclasses.
Following the terminology from that work, this mapping is: \allthree{} $\to$
\textsc{DownUp-36}, \allfour{} $\to$ \textsc{UpDown-127}, \living{} $\to$
\textsc{DownUp-753} (restricted to ``living'' nodes), and \nonliving{} $\to$
\textsc{DownUp-345} (restricted to ``non-living'' nodes).

\clearpage
\subsection{Evaluating model performance}
\label{app:eval_setup}

\subsubsection{Model architectures and training}
\label{app:models}
The model architectures used in our analysis are in
Table~\ref{tab:models} for which we used standard implementations
from the PyTorch library 
(\url{https://pytorch.org/docs/stable/torchvision/models.html}).
For training, we use a batch size of 128, weight decay of $10^{-4}$, 
and learning rates listed in Table~\ref{tab:models}.
Models were trained until convergence.
On \allthree{} and \allfour{}, this required a total of 300 epochs, 
with 10-fold drops in learning rate every 100 epochs, while on
\living and \nonliving, models a total of 450 epochs, with 10-fold learning rate
drops every 150 epochs.
For adapting models, we retrained the last
(fully-connected) layer on the train split of the target domain, starting from
the parameters of the source-trained model.
We trained that layer using SGD with a batch size of 128 for
40,000 steps and chose the best learning rate out of 
$[0.01, 0.1, 0.25, 0.5, 1.0, 2.0, 3.0, 5.0, 7.0, 8.0, 10.0, 11.0, 12.0]$, based
on test accuracy.

\begin{table}[!h]
\begin{center}
	\begin{tabular}{lcc}
		\toprule
		\textbf{Model} & \phantom{x} & \textbf{Learning Rate}  \\
		\midrule
		\texttt{alexnet} && 0.01 \\ 
		\texttt{vgg11} && 0.01  \\ 
		\texttt{resnet18} && 0.1 \\ 
		\texttt{resnet34} && 0.1 \\ 
		\texttt{resnet50} && 0.1 \\ 
		\texttt{densenet121} && 0.1  \\ 
		\bottomrule
	\end{tabular}
\end{center}
	\caption{Models used in our analysis.} 
	\label{tab:models}
\end{table}

\subsubsection{Model pairwise accuracy}
\label{app:model_pairwise}
In order to make a fair comparison between the performance of models and human 
annotators on the \Breeds{} tasks, we evaluate model accuracy on
pairs of superclasses. On images from that pair,
we determine the model prediction to be the superclass for 
which the model's predicted probability is higher. A prediction is deemed correct if it 
matches the superclass label for the image. Repeating this process over random pairs 
of superclasses allows us to estimate model accuracy on the average-case binary
classification task.

\subsubsection{Robustness interventions}
\label{app:robustness}
For model training, we use the hyperparameters provided in 
Appendix~\ref{app:models}.
Additional intervention-specific hyperparameters are listed in Appendix 
Table~\ref{tab:ri_hyperparams}.   Due to computational 
constraints, we trained a restricted set of model architectures with robustness 
interventions---ResNet-18 and ResNet-50 for adversarial training, and ResNet-18 
and ResNet-34 for all others.
Adversarial training was implemented using the \texttt{robustness}
library,\footnote{\url{https://github.com/MadryLab/robustness}} while random
erasing using the PyTorch
\texttt{transforms}.\footnote{\url{https://pytorch.org/docs/stable/torchvision/transforms.html}}

\begin{table}[!h]
	\begin{center}
	\begin{minipage}{0.3\textwidth}
		\begin{tabular}{ccc}
			\toprule
			\textbf{Eps} & \textbf{Step size} & \textbf{\#Steps} \\
			\midrule
      0.5 & 0.4 & 3 \\ 
			1 & 0.8 & 3 \\ 
			\bottomrule
		\end{tabular}
	
    \caption*{(a) PGD-training~\citep{madry2018towards}}
\end{minipage}	
\hfil
\begin{minipage}{0.2\textwidth}
			\begin{tabular}{cc}
		\toprule
		\textbf{Mean}  &	\textbf{StdDev}  \\
		\midrule
        0  & 0.2  \\ 
		\bottomrule
	\end{tabular}
	\caption*{(b) Gaussian noise} 
\end{minipage}	
\hfil
\begin{minipage}{0.4\textwidth}
			\begin{tabular}{ccc}
		\toprule
        \textbf{Probability}  &	\textbf{Scale} & \textbf{Ratio} \\
		\midrule
        0.5 & 0.02 - 0.33 & 0.3 - 3.3 \\
		\bottomrule
	\end{tabular}
	\caption*{(c) Random erasing} 
\end{minipage}	
	\end{center}
	\caption{Additional hyperparameters for robustness interventions.} 
\label{tab:ri_hyperparams}
\end{table}

\clearpage
\section{Additional Experimental Results}
\label{app:res}

\subsection{Human Baselines for \Breeds{} Tasks}
\label{app:res_human}
\begin{figure}[!h]
	
	\begin{subfigure}{1\textwidth}
		\centering
			\includegraphics[width=0.82\textwidth]{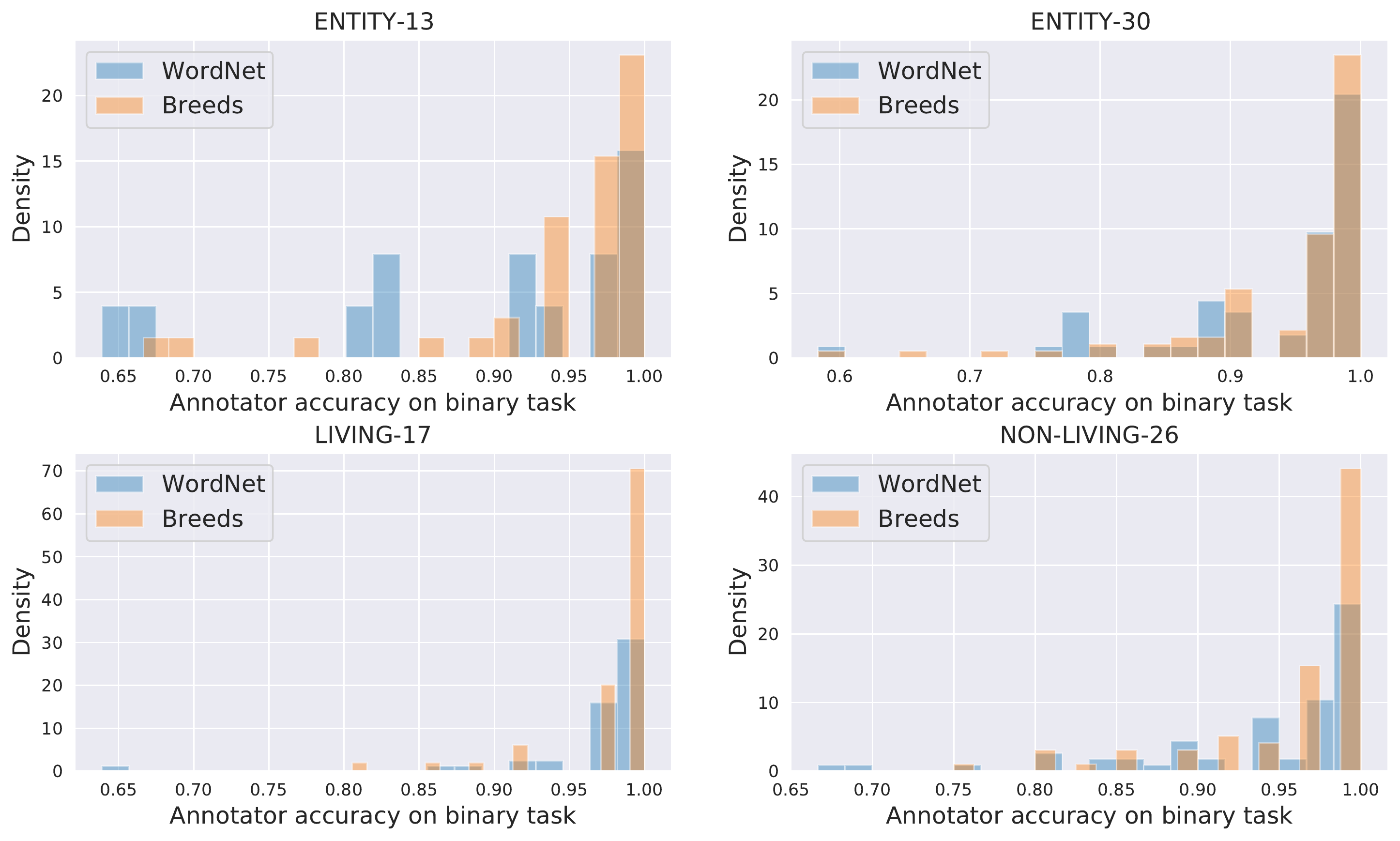}
			\caption{Source domain (no subpopulation shift)}
			\label{fig:human_pairwise_s}
	\end{subfigure}
\begin{subfigure}{1\textwidth}
	\centering
	\includegraphics[width=0.82\textwidth]{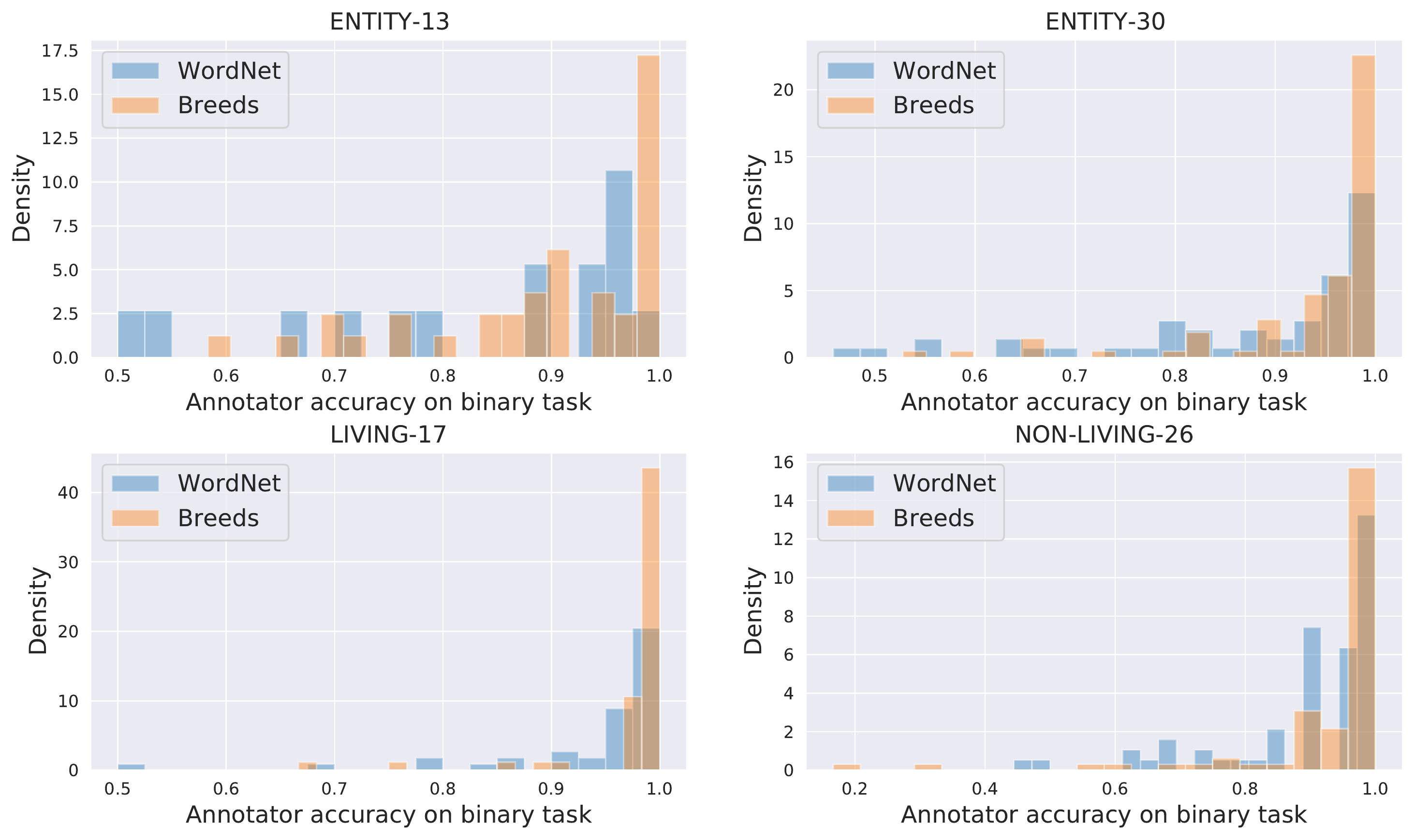}
	\caption{Target domain (with subpopulation shift)}
	\label{fig:human_pairwise_t}
\end{subfigure}
	\caption{Distribution of annotator accuracy over pairwise superclass classification 
		tasks. We observe that human annotators consistently perform 
		better 
		on tasks 
		constructed using our modified ImageNet class 
		hierarchy (i.e., \Breeds{}) as opposed to those obtained directly from 
		WordNet.}

\end{figure}

In Section~\ref{sec:humans}, we evaluate human performance on binary versions of 
our \Breeds{} tasks. Appendix Figures~\ref{fig:human_pairwise_s} 
and~\ref{fig:human_pairwise_t} show the distribution of annotator accuracy over 
different pairs of superclasses for test data sampled from the source and target 
domains respectively.

\clearpage

\subsection{Model Evaluation}
\label{app:res_eval}
\begin{figure}[!h]
	\centering
	\includegraphics[width=1\textwidth]{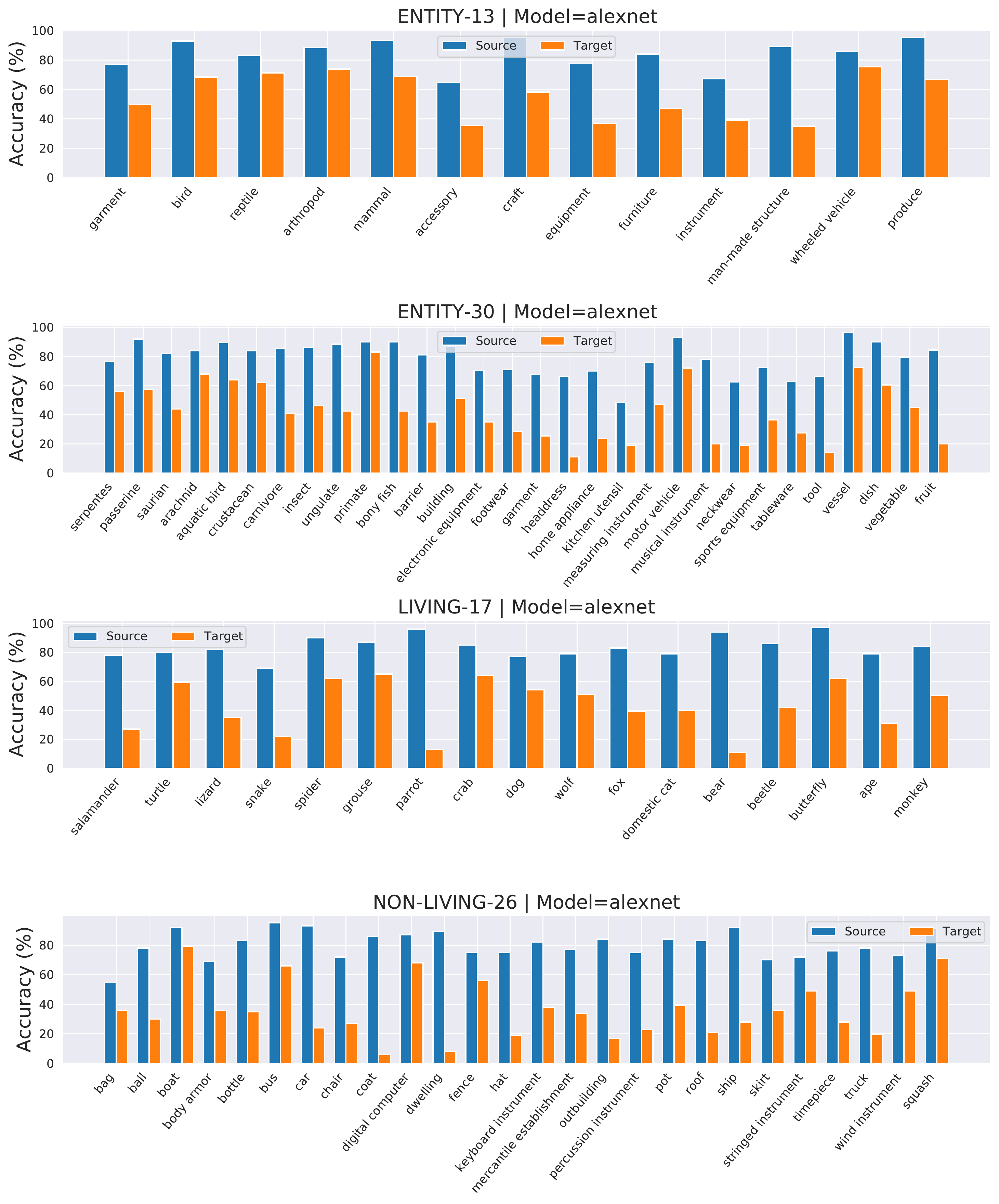}
	\caption{Per-class source and target accuracies for AlexNet on \Breeds{}
  tasks.}
	\label{fig:perclass_a}
\end{figure}

\begin{figure}[!h]
	\centering
	\includegraphics[width=1\textwidth]{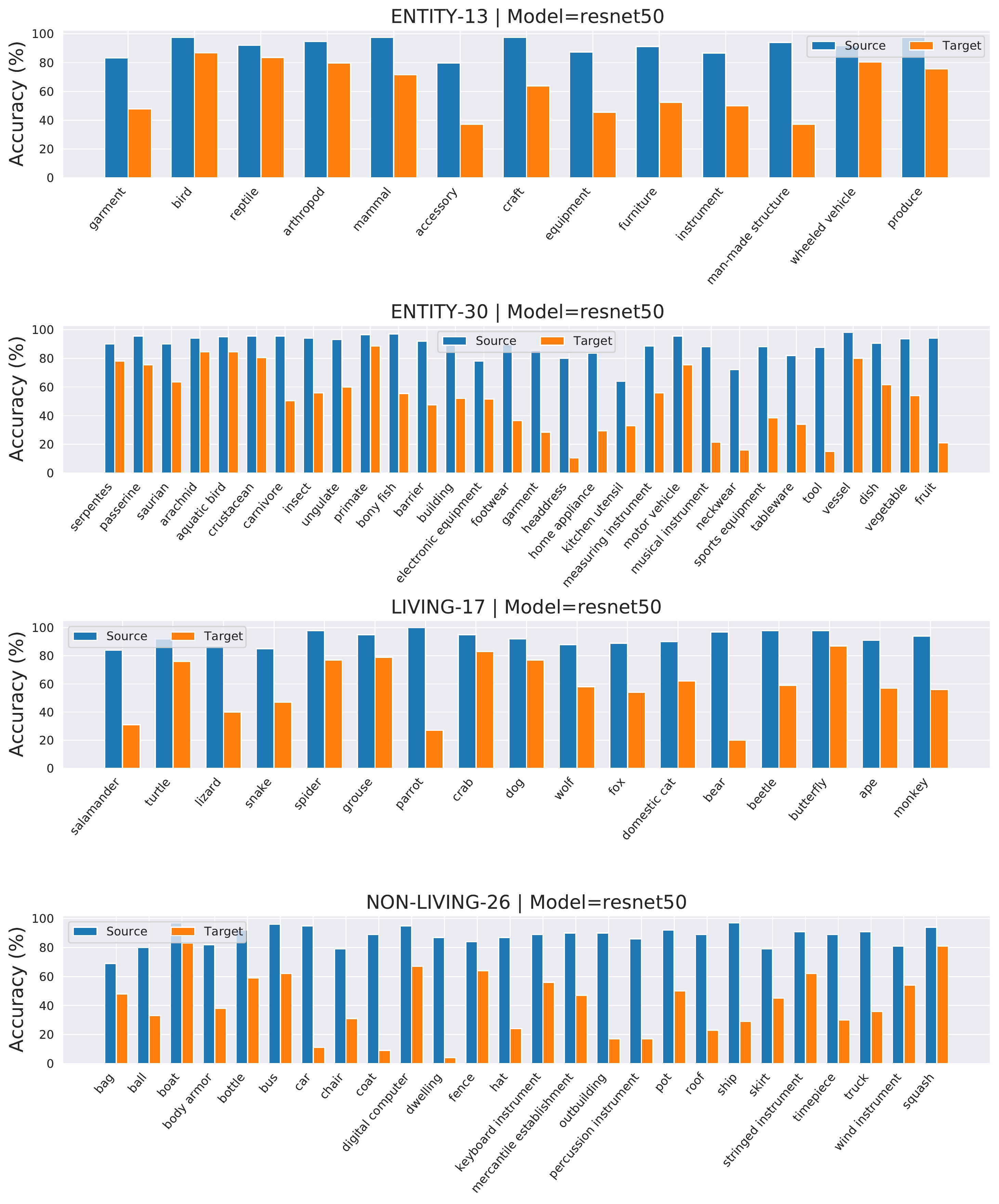}
	\caption{Per-class source and target accuracies for ResNet-50 on \Breeds{}
    tasks.}
	\label{fig:perclass_r50}
\end{figure}

\begin{figure}[!h]
	\centering
	\includegraphics[width=1\textwidth]{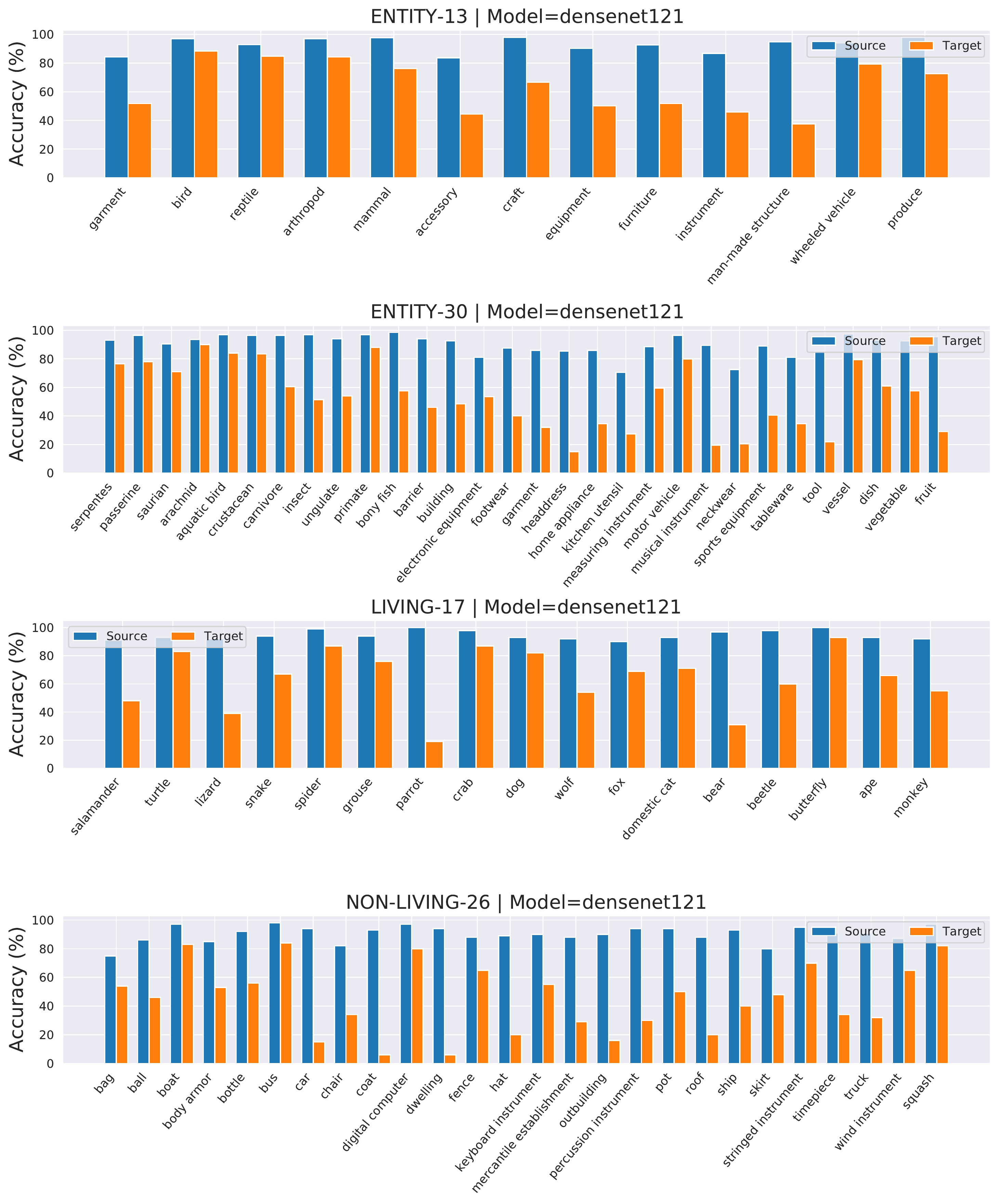}
	\caption{Per-class source and target accuracies for DenseNet-121 on \Breeds{}
    tasks.}
	\label{fig:perclass_d121}
\end{figure}

In Figures~\ref{fig:perclass_a}-~\ref{fig:perclass_d121}, we visualize model 
performance over \Breeds{} superclasses for different model architectures. We observe in 
general that models perform fairly uniformly over classes when the test data is drawn 
from the source domain. This indicates that the tasks are well-calibrated---the 
various superclasses are of comparable difficulty. At the same time, we see that 
model robustness to subpopulation shift, i.e., drop in accuracy on the target domain, 
varies widely over superclasses. This could be either due to some superclasses
being broader by construction or due to models being more sensitive to
subpopulation shift for some classes.


\clearpage
\subsubsection{Effect of different splits}
\label{app:res_splits}
\label{app:goodbad}

As described in Section~\ref{sec:breeds}, to create \Breeds{} tasks, we first identify 
a set of relevant superclasses (at the chosen depth in the hierarchy), and then 
partition their subpopulations between the source and target domains. For all the 
tasks listed in Table~\ref{tab:benchmarks}, the superclasses are balanced---each of 
them comprise the same number of subpopulations. To ensure this is the case, the 
desired number of subpopulations is chosen among all superclass subpopulations at 
random. These subpopulations are then randomly split between the source and target 
domains.

\begin{figure}[!h]
	\centering
	\includegraphics[width=0.8\textwidth]{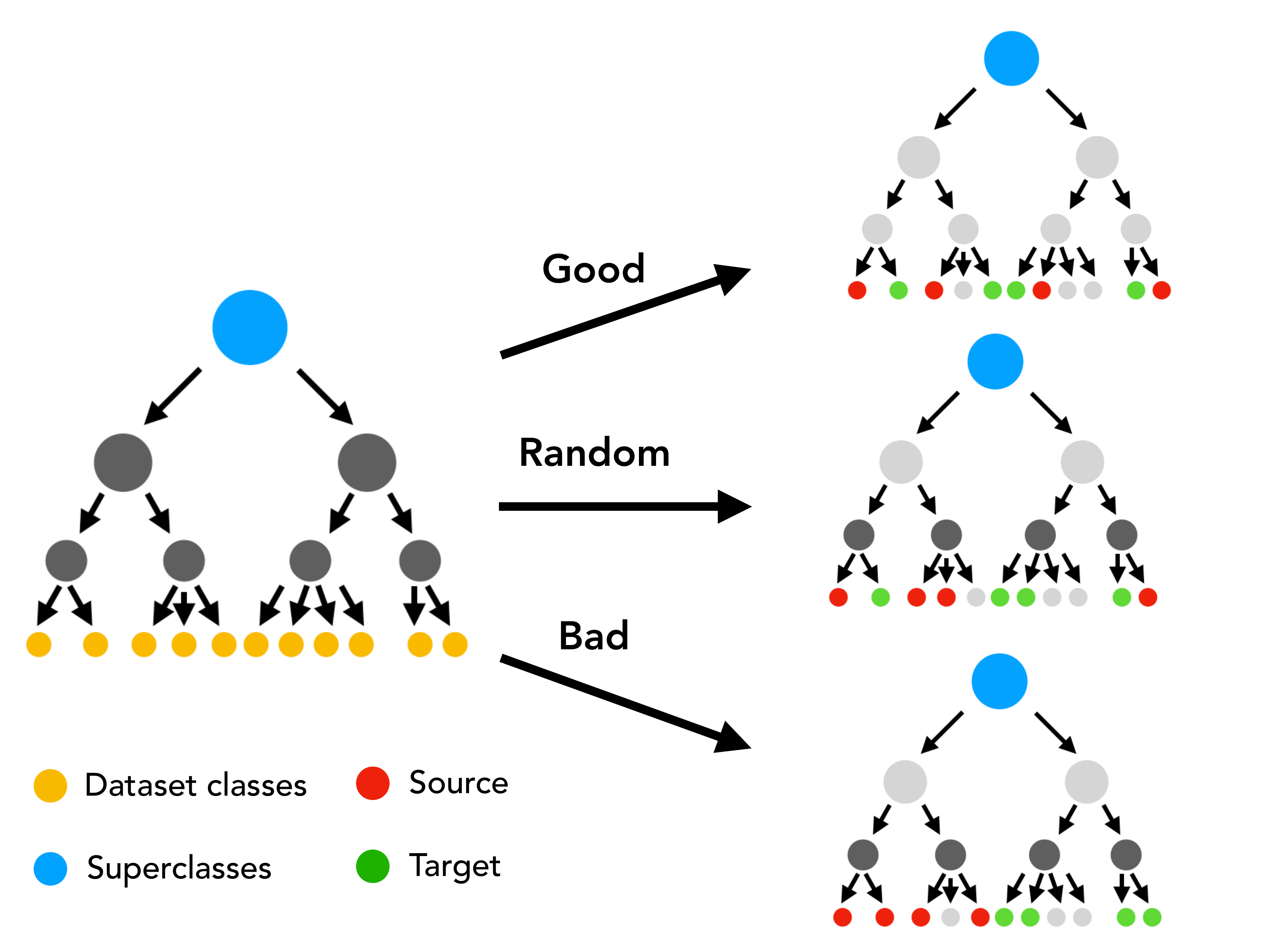}
	\caption{Different ways to partition the subpopulations of a given superclass into 
		the source and target domains. Depending on how closely related the 
		subpopulations in the two domain are, we can construct splits that are more/less 
		adversarial.}
	\label{fig:splits_diag}
\end{figure}

Instead of randomly partitioning subpopultions (of a given superclass) between the 
two domains, we could instead craft partitions to be more/less adversarial as 
illustrated in Figure~\ref{fig:splits_diag}. Specifically, we could control how similar 
the subpopulations in the target domain are to those in the source domain. For 
instance, a split would be less adversarial (\emph{good}) if subpopulations in the 
source and target domain share a common parent. On the other hand, we could make 
a split more adversarial (\emph{bad}) by ensuring a greater degree of separation (in 
terms of distance in the hierarchy) between the source and target domain 
subpopulations.

\begin{figure}[!h]
	\begin{subfigure}{1.0\textwidth}
	\centering
	\includegraphics[width=1\textwidth]{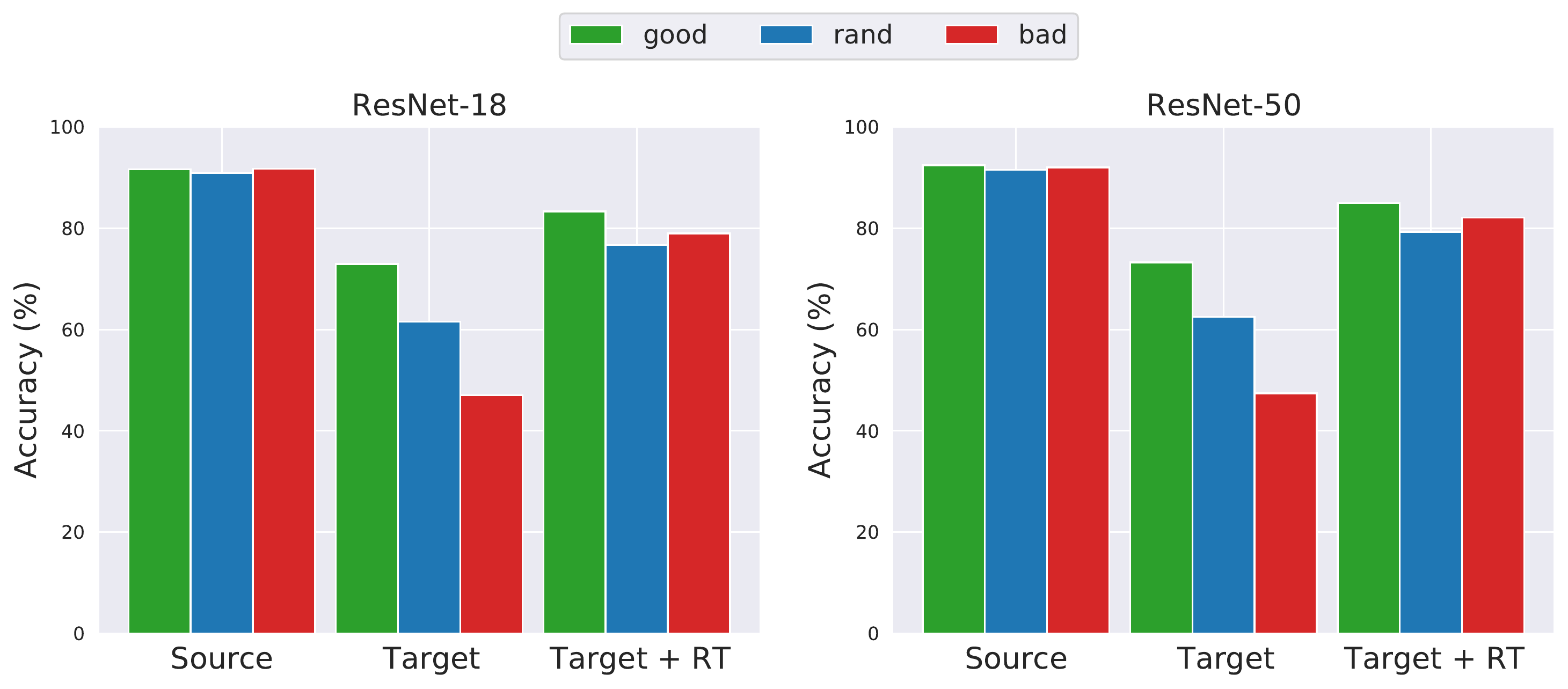}
	\caption{\allthree{} task}
	\label{fig:all3_splits}
	\end{subfigure}
	\begin{subfigure}{1.0\textwidth}
	\centering
	\includegraphics[width=1\textwidth]{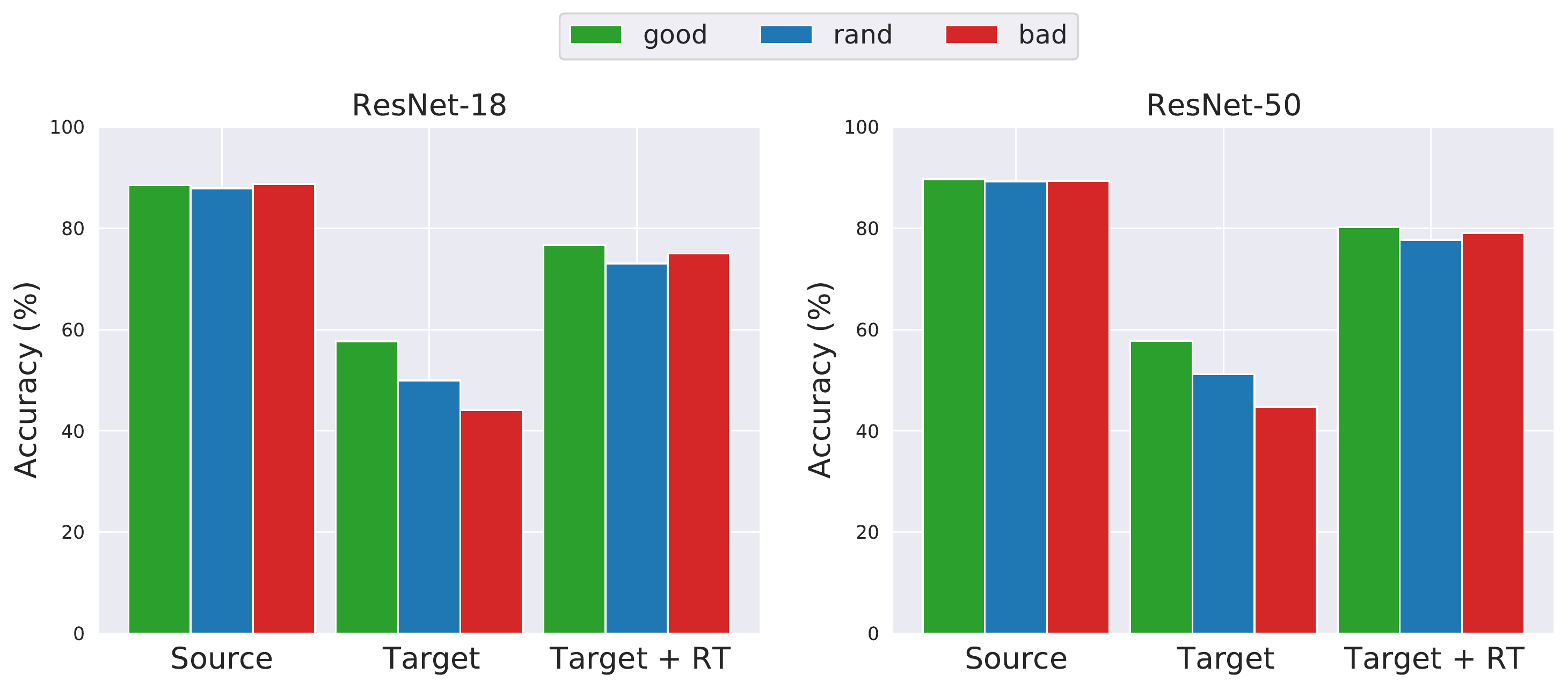}
	\caption{\allfour{} task}
	\label{fig:all4_splits}
	\end{subfigure}
	\caption{Model robustness as a function of the nature of subpopulation shift within 
	specific \Breeds{} tasks. We vary how the underlying 
	subpopulations of each superclass are split between the source and target 
	domain---we 
	compare random splits (used in the majority of our analysis), to ones that are more 
	(\emph{bad})
	or less adversarial (\emph{good}).
	When models are tested on samples from the source domain, they perform equally 
	well across different splits, as one might expect.
	However, under subpopulation shift (i.e., on samples from the target domain), 
	model robustness varies drastically, and is considerably worse when the split is  
	more adversarial.
	Yet, for all the splits, models have comparable target accuracy 
    after retraining their final layer.
}
\label{fig:all_splits}
\end{figure}

 We now evaluate model performance under such variations in the nature of the splits 
 themselves---see Figure~\ref{fig:all_splits}. 
As expected, models perform comparably well on test data from the source domain, 
independent of the how the subpopulations are partitioned into the two domains. 
However, model robustness to subpopulation 
shift varies considerably based on the nature of the split---it is lowest for the most 
adversarially chosen split. 
Finally, we observe that retraining the linear layer 
on data from the target domain recovers a considerable fraction of the accuracy drop 
in all cases---indicating that even for the more adversarial splits, models do learn 
features that transfer well to unknown subpopulations. 

\clearpage
\subsubsection{Robustness Interventions}
\label{app:res_int}
In Tables~\ref{tab:adv_app} and~\ref{tab:other_rob_app}, we present the raw 
accuracies of models trained using various train-time robustness interventions.
\begin{table}[!h]
  \setlength{\tabcolsep}{1.5em}
	\centering
	\renewcommand{\arraystretch}{1.05}
	\begin{tabular}{llccc}
		\toprule
		\multicolumn{5}{c}{ResNet-18} \\
		\midrule
    \multirow{2}{*}{Task} & \multirow{2}{*}{$\eps$} & \multicolumn{3}{c}{Accuracy (\%)} \\
    & &  Source & Target & Target-RT \\
		\midrule
    \multirow{3}{*}{\allthree} &    0 &  \textbf{90.91 $\pm$  0.73} &
      \textbf{61.52 $\pm$  1.23} &  \textbf{76.71 $\pm$  1.09} \\
    &  0.5 &  89.23 $\pm$  0.80 &  \textbf{61.10 $\pm$  1.23} &  74.92 $\pm$  1.04 \\
 &  1.0 &  88.45 $\pm$  0.81 &  58.53 $\pm$  1.26 &  73.35 $\pm$  1.11 \\
		\midrule
     \multirow{3}{*}{\allfour} &    0 &  \textbf{87.88 $\pm$  0.89} &
     \textbf{49.96 $\pm$  1.31} &  \textbf{73.05 $\pm$  1.17}
     \\
    &  0.5 &  85.68 $\pm$  0.91 &  \textbf{48.93 $\pm$  1.34} &  71.34 $\pm$  1.14 
\\
 &  1.0 &  84.23 $\pm$  0.91 &  47.66 $\pm$  1.23 &  70.27 $\pm$  1.17 \\
		\midrule
    \multirow{3}{*}{\living} &    0 &  \textbf{92.01 $\pm$  1.30} & 
      \textbf{58.21 $\pm$  2.32} &  \textbf{83.38 $\pm$  1.79} \\
    &  0.5 &  90.35 $\pm$  1.35 &  55.79 $\pm$  2.44 & \textbf{83.00 $\pm$ 1.89} \\
 &  1.0 &  88.56 $\pm$  1.50 &  53.89 $\pm$  2.36 &  80.90 $\pm$  1.92 \\
		\midrule
    \multirow{3}{*}{\nonliving} &    0 &  \textbf{88.09 $\pm$  1.28} &  
      \textbf{41.87 $\pm$  2.01} & \textbf{73.52 $\pm$  1.71} \\
    &  0.5 &  86.28 $\pm$  1.32 &  \textbf{41.02 $\pm$  1.91} &
    \textbf{72.41 $\pm$  1.71} \\
    &  1.0 &  85.19 $\pm$  1.38 &  \textbf{40.23 $\pm$  1.92} &  70.61 $\pm$  
1.73 \\
		\bottomrule 
	\end{tabular}
	\begin{tabular}{llccc}
		\multicolumn{5}{c}{} \\
	\toprule
		\multicolumn{5}{c}{ResNet-50} \\
		\midrule
    \multirow{2}{*}{Task} & \multirow{2}{*}{$\eps$} & \multicolumn{3}{c}{Accuracy (\%)} \\
    & &  Source & Target & Target-RT \\
	\midrule
    \multirow{3}{*}{\allthree} &    0 &  \textbf{91.54 $\pm$  0.64} &
    \textbf{62.48 $\pm$  1.16} &  \textbf{79.32 $\pm$  1.01 } \\
    &  0.5 &  89.87 $\pm$  0.80 &  \textbf{63.01 $\pm$  1.15} & 
     \textbf{80.14 $\pm$  1.00} \\
 &  1.0 &  89.71 $\pm$  0.74 &  61.21 $\pm$  1.22 &  78.58 $\pm$  0.98 \\
	\midrule
    \multirow{3}{*}{\allfour} &    0 &  \textbf{89.26 $\pm$  0.78} &
    \textbf{51.18 $\pm$  1.24} &  77.60 $\pm$  1.17 \\

    &  0.5 &  87.51 $\pm$  0.88 &  \textbf{50.72 $\pm$  1.28} & 
      \textbf{78.92 $\pm$  1.06} \\
    &  1.0 &  86.63 $\pm$  0.88 &  \textbf{50.99 $\pm$  1.27} & 
      \textbf{78.63 $\pm$  1.03} \\
	\midrule
    \multirow{3}{*}{\living}  &    0 &  \textbf{92.40 $\pm$  1.28} &
    \textbf{58.22 $\pm$  2.42} &  \textbf{85.96 $\pm$  1.72} \\
    &  0.5 &  90.79 $\pm$  1.55 &  \textbf{55.97 $\pm$  2.38} & 
      \textbf{87.22 $\pm$  1.66} \\
    &  1.0 &  89.64 $\pm$  1.47 &  54.64 $\pm$  2.48 &  \textbf{85.63 $\pm$ 1.73} \\
	\midrule
    \multirow{3}{*}{\nonliving}  &    0 &  \textbf{88.13 $\pm$  1.30} & 
      \textbf{41.82 $\pm$  1.86} &  76.58 $\pm$  1.69 \\
    &  0.5 &  \textbf{88.20 $\pm$  1.20} &  \textbf{42.57 $\pm$  2.03} &
      \textbf{78.84 $\pm$  1.62} \\
    &  1.0 &  86.17 $\pm$  1.36 &  \textbf{41.69 $\pm$  1.96} &  76.16 $\pm$  
1.61 \\
	\bottomrule
\end{tabular}
\vspace{1em}
	\caption{Effect of adversarial training on model robustness to subpopulation 
	shift. All models are trained on samples from the source domain---either 
	using standard 
	training ($\eps=0.0$) or using adversarial training. Models are then 
	evaluated in terms of: (a) source accuracy, (b) target accuracy and (c) target 
	accuracy after retraining the linear layer of the model with data from the 
  target domain. Confidence intervals (95\%) obtained via bootstrapping. Maximum
  task accuracy over $\eps$ (taking into account confidence interval) shown in bold.}
	\label{tab:adv_app}
\end{table}

\begin{table}[!h]
	\centering
	\renewcommand{\arraystretch}{1.05}
	\begin{tabular}{llccc}
		\multicolumn{5}{c}{} \\
	\toprule
		\multicolumn{5}{c}{ResNet-18} \\
		\midrule
    \multirow{2}{*}{Task} & \multirow{2}{*}{Intervention} & \multicolumn{3}{c}{Accuracy (\%)} \\
    & &  Source & Target & Target-RT \\
		\midrule
    \multirow{4}{*}{\allthree}  &       Standard & 
      \textbf{90.91 $\pm$  0.73} &  \textbf{61.52 $\pm$  1.23 } &
      76.71 $\pm$  1.09 \\
    &     Erase  Noise&  \textbf{91.01 $\pm$  0.68} & \textbf{62.79 $\pm$  1.27}
    &  \textbf{78.10 $\pm$  1.09} \\
 &  Gaussian Noise &  77.00 $\pm$  1.04 &  47.90 $\pm$  1.21 &  70.37 
$\pm$  1.17 \\
&  Stylized ImageNet &  76.85 $\pm$  1.00 &  50.18 $\pm$  1.21 &  65.91 
$\pm$  
1.17 \\
	\midrule
    \multirow{4}{*}{\allfour} &       Standard &  \textbf{87.88 $\pm$  0.89} & 
    \textbf{49.96 $\pm$  1.31 } &  73.05 $\pm$  1.17 \\
    &     Erase Noise &  \textbf{88.09 $\pm$  0.80} &  \textbf{49.98 $\pm$  1.31}
      & \textbf{74.27 $\pm$  1.15} \\
&  Gaussian Noise&  74.12 $\pm$  1.16 &  35.79 $\pm$  1.21 &  65.62 
$\pm$  1.28 \\
 &  Stylized ImageNet &  70.96 $\pm$  1.16 &  37.67 $\pm$  1.21 &  60.45 
$\pm$  1.22 \\
	\midrule
    \multirow{4}{*}{\living} &       Standard &  \textbf{92.01 $\pm$  1.30} &
    \textbf{58.21 $\pm$  2.32} &  \textbf{83.38 $\pm$  1.79} \\
    &     Erase Noise &  \textbf{93.09 $\pm$  1.27} &  \textbf{59.60 $\pm$ 2.40}
    &  \textbf{85.12 $\pm$  1.71} \\
&  Gaussian Noise &  80.13 $\pm$  1.99 &  46.16 $\pm$  2.57 &  77.31 
$\pm$  
2.08 \\
 &  Stylized ImageNet &  79.21 $\pm$  1.85 &  43.96 $\pm$  2.38 &  72.74 
$\pm$  
2.09 \\
	\midrule
\multirow{4}{*}{\nonliving} &   
    Standard &  \textbf{88.09 $\pm$  1.28} &  \textbf{41.87 $\pm$  2.01 }
      &  \textbf{73.52 $\pm$  1.71} \\
    &     Erase Noise &  \textbf{88.68 $\pm$  1.18} &  \textbf{43.17 $\pm$ 2.10}
    &  \textbf{73.91 $\pm$  1.78} \\
&  Gaussian Noise &  78.14 $\pm$  1.60 &  35.13 $\pm$  1.94 &  67.79 
$\pm$  1.79 \\
 &  Stylized ImageNet &  71.43 $\pm$  1.73 &  30.56 $\pm$  1.75 &  61.83 
$\pm$  1.98 \\
		\bottomrule 
	\end{tabular}
	\begin{tabular}{llccc}
		\multicolumn{5}{c}{} \\
	\toprule
		\multicolumn{5}{c}{ResNet-34} \\
		\midrule
    \multirow{2}{*}{Task} & \multirow{2}{*}{Intervention} & \multicolumn{3}{c}{Accuracy (\%)} \\
    & &  Source & Target & Target-RT \\
		\midrule
    \multirow{4}{*}{\allthree} &       Standard &  \textbf{91.75 $\pm$  0.70}
    & \textbf{ 63.45 $\pm$  1.13 } &  \textbf{ 78.07 $\pm$  1.02} \\
    &     Erase Noise &  \textbf{91.76 $\pm$  0.70}
      &  \textbf{62.71 $\pm$ 1.25} &  \textbf{77.43 $\pm$  1.06} \\
 &  Gaussian Noise &  81.60 $\pm$  0.97 &  50.69 $\pm$  1.28 &  71.50 
$\pm$  1.13 \\
&  Stylized ImageNet &  78.66 $\pm$  0.94 &  51.05 $\pm$  1.30 &  67.38 
$\pm$  1.16 \\
\midrule
    \multirow{4}{*}{\allfour} &       Standard &  \textbf{88.81 $\pm$  0.81} & 
    \textbf{51.68 $\pm$  1.28 } &  \textbf{75.12 $\pm$  1.11} \\
    &     Erase Noise &  \textbf{89.07 $\pm$  0.82} &  \textbf{51.04 $\pm$
    1.27} &  \textbf{74.88 $\pm$  1.08} \\
&  Gaussian Noise &  75.05 $\pm$  1.11 &  38.31 $\pm$  1.26 &  67.47 
$\pm$  1.22 \\
 &  Stylized ImageNet &  72.51 $\pm$  1.10 &  38.98 $\pm$  1.22 &  61.65 
$\pm$  1.25 \\
\midrule
    \multirow{4}{*}{\living} &       Standard &  \textbf{92.83 $\pm$  1.19}
      &  59.74 $\pm$  2.27 &  \textbf{85.46 $\pm$  1.83} \\
    &     Erase Noise &  \textbf{92.96 $\pm$  1.32} &  \textbf{61.13 $\pm$
    2.30} &  \textbf{85.66 $\pm$  1.78} \\
&  Gaussian Noise &  84.06 $\pm$  1.71 &  48.38 $\pm$  2.44 &  78.79 
$\pm$  1.91 \\
 &  Stylized ImageNet &  80.94 $\pm$  2.00 &  44.16 $\pm$  2.43 &  72.77 
$\pm$  2.18 \\
\midrule
    \multirow{4}{*}{\nonliving} &       Standard &  \textbf{89.64 $\pm$  1.17}
      &  \textbf{43.03 $\pm$  1.99 } &  \textbf{74.99 $\pm$  1.66} \\
    &     Erase Noise &  \textbf{89.62 $\pm$   1.31} &  \textbf{43.53 $\pm$   
    1.89} &  \textbf{75.04 
    $\pm$   1.70} \\
 &  Gaussian Noise &  79.26 $\pm$   1.61 &  34.89 $\pm$   1.91 &  68.07 
 $\pm$   1.78
 \\
 &  Stylized ImageNet&  71.49 $\pm$  1.65 &  31.10 $\pm$  1.80 &  62.94 
$\pm$  1.90 \\
		\bottomrule
	\end{tabular}
	\vspace{1em}
	\caption{Effect of various train-time interventions on model robustness to 
	subpopulation 
	shift. All models are trained on samples from the the source domain. Models 
	are then 
	evaluated in terms of: (a) source accuracy, (b) target accuracy and (c) target 
	accuracy after retraining the linear layer of the model with data from the 
  target domain. Confidence intervals (95\%) obtained via bootstrapping. Maximum
  task accuracy over $\eps$ (taking into account confidence interval) shown in bold.}
	\label{tab:other_rob_app}
\end{table}

\end{document}